\documentclass[10pt,twocolumn,letterpaper]{article}

\usepackage{iccv}
\usepackage{times}
\usepackage{epsfig}
\usepackage{graphicx}
\usepackage{amsmath}
\usepackage{amssymb}
\usepackage{algorithm}
\usepackage{algpseudocode}

\usepackage[accsupp]{axessibility}
\usepackage{multirow}
\usepackage{booktabs}
\usepackage{bbm}
\usepackage{caption}
\usepackage{subcaption}
\usepackage{placeins}
\usepackage{rotating}
\usepackage{longtable}
\usepackage{mdframed}
\usepackage{appendix}
\usepackage{xcolor}
\newcommand{\stdev}[1]{\textcolor{gray}{\tiny{$\pm$#1}}}
\usepackage[pagebackref=true,breaklinks=true,letterpaper=true,colorlinks,bookmarks=false]{hyperref}
\usepackage[sectionbib]{chapterbib}
\usepackage{chapterbib}

\iccvfinalcopy

\def\iccvPaperID{9882}

\ificcvfinal\fi

\begin{document}

\title{Distribution-Aware Prompt Tuning for Vision-Language Models}

\author{Eulrang Cho\thanks{Equal contribution.} \quad \quad Jooyeon Kim\protect\footnotemark[\arabic{footnote}] \quad \quad Hyunwoo J. Kim\thanks{Corresponding author.}\vspace{0.4cm}\\
Department of Computer Science and Engineering, Korea University\\
{\tt\small \{ercho, parang, hyunwoojkim\}@korea.ac.kr}
}

\maketitle
\ificcvfinal\thispagestyle{empty}\fi

\begin{abstract} 
   Pre-trained vision-language models (VLMs) have shown impressive performance on various downstream tasks by utilizing knowledge learned from large data. 
In general, the performance of VLMs on target tasks can be further improved by prompt tuning, which adds context to the input image or text. 
By leveraging data from target tasks, various prompt-tuning methods have been studied in the literature. 
A key to prompt tuning is the feature space alignment between two modalities via learnable vectors with model parameters fixed.
We observed that the alignment becomes more effective when embeddings of each modality are `well-arranged' in the latent space.
Inspired by this observation, we proposed distribution-aware prompt tuning (DAPT) for vision-language models, which is simple yet effective.
Specifically, the prompts are learned by maximizing inter-dispersion, the distance between classes, as well as minimizing the intra-dispersion measured by the distance between embeddings from the same class. 
Our extensive experiments on 11 benchmark datasets demonstrate that our method significantly improves generalizability.
The code is available at \href{https://github.com/mlvlab/DAPT}{https://github.com/mlvlab/DAPT}.
\end{abstract}

\section{Introduction}
\begin{figure}
    \centering
    \begin{tabular}{ccc}
    \multirow{-8}{*}{\rotatebox[origin=c]{90}{Text}} & \subfloat[Zero-shot CLIP]{\includegraphics[width=.2\textwidth]{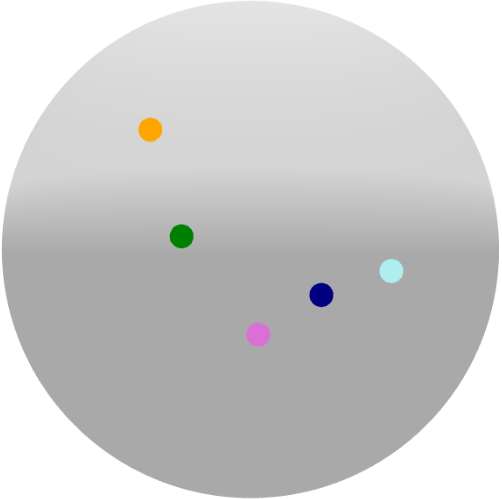}
    \label{fig:tsne_text_zs}}
    & \subfloat[DAPT]{\includegraphics[width=.2\textwidth]{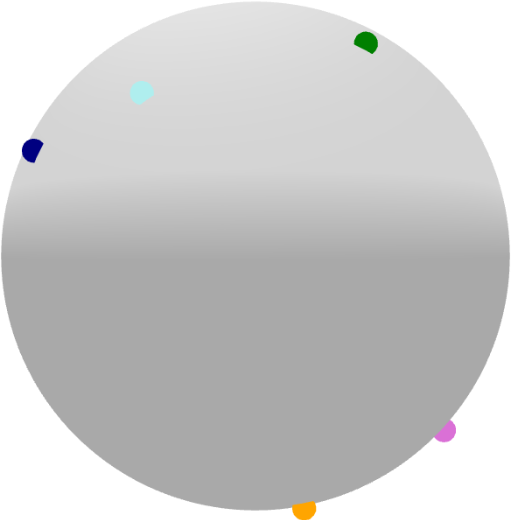}
    \label{fig:tsne_text}} \\
    \\
    \multirow{-7.5}{*}{\rotatebox[origin=l]{90}{Image}} & 
    \subfloat[Zero-shot CLIP]{\includegraphics[width=.2\textwidth]{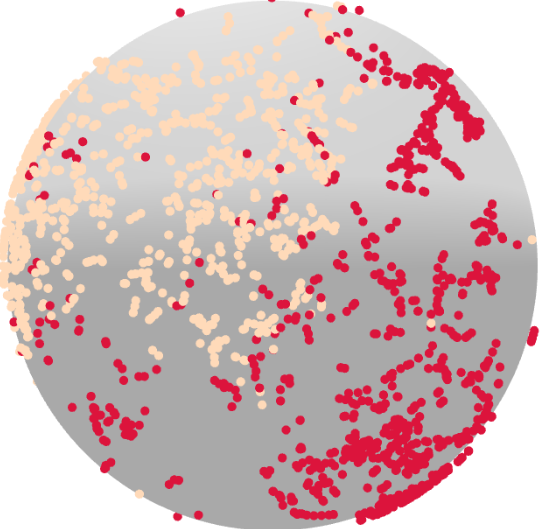}
    \label{fig:tsne_image_zs}} 
    & \subfloat[DAPT]{\includegraphics[width=.2\textwidth]{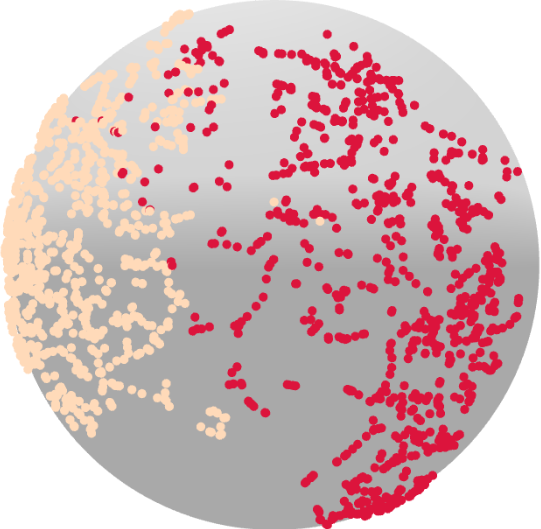}
    \label{fig:tsne_image_text}}  \\

    \end{tabular}
    \caption{
    Points in (a) and (b) denote normalized t-SNE embeddings of text features from OxfordPets~\cite{parkhi2012cats}. In addition, points on (c) and (d) represent the image features' t-SNE embeddings from EuroSAT~\cite{helber2019eurosat}. As shown in (a) and (b), zero-shot CLIP has small distances between text embeddings of class labels, but its image embeddings do not cluster well. However, prompt-tuning with DAPT leads to more evenly spaced text embeddings and better clusterings of image embeddings within the same class. 
    }
    \label{fig:tsne}
\end{figure}
In recent years, pre-trained vision-language models (VLMs) have shown great success in a wide range of applications in computer vision such as 
image classification \cite{radford2021learning,singh2022flava}, object detection \cite{Du_2022_CVPR,gu2022openvocabulary,zhong2022regionclip}, captioning \cite{li2023blip2,mokady2021clipcap,zhang2023llamaadapter}, and visual question answering (VQA) \cite{garcia2020knowit}.
Notably, VLMs have shown promising generalization power and transferability in various downstream tasks. 
For instance, VLMs such as CLIP~\cite{radford2021learning} and ALIGN~\cite{jia2021scaling} show outstanding performance in
zero-shot and few-shot learning.
These models opened the door for zero-shot image classification and zero-shot object detection.
To further improve the pre-trained models' zero-shot generalization ability, \textit{prompting} has been proposed. 
For instance, in image classification, CLIP~\cite{radford2021learning} suggests using a context text ``{\tt A photo of a }'' in front of the class label {\tt [CLASS]} to obtain text embeddings for target classes. 

Prompting is an emerging research topic due to several advantages over fine-tuning, which is a conventional approach to utilize pre-trained deep-learning models.
For a pre-trained VLM, fine-tuning is often practically challenging due to the large number of model parameters.
In addition, fine-tuning the entire VLM often results in overfitting due to a small amount of target domain data. 
Zhou~\etal~\cite{zhou2022coop} have shown that more powerful context strings (\textit{hard prompts}) exist. 
However, manually finding better hard prompts (prompt engineering) is time-consuming and suboptimal. 
So, after that, a line of works has proposed prompt-tuning that optimizes \textit{soft prompts}, learnable vectors~\cite{jia2022visual,lester-etal-2021-power,zhou2022coop}. 
The learnable vectors are concatenated with other inputs and numerically optimized by backpropagation while the pre-trained VLM models parameters are fixed.

The prompt tuning can be viewed as the alignment between two latent spaces of text and image. 
Figure~\ref{fig:tsne} shows that each latent space of CLIP is not suitable for feature alignment.
The text embeddings of target classes obtained from the original CLIP in Figure~\ref{fig:tsne_text_zs} are gathered nearby, which potentially leads to misclassification to close classes. 
In addition, the original CLIP's visual embeddings in Figure~\ref{fig:tsne_image_zs} are widely spread, and some regions are overlapped. 
To address this problem, we propose a prompt tuning method, \textbf{DAPT}\footnote{\textbf{D}istribution-\textbf{A}ware \textbf{P}rompt \textbf{T}uning}, that optimizes the distribution of embeddings for each modality for better feature alignment.

DAPT learns vectors (\ie, soft prompts) for both text and image encoders with additional loss terms\ - inter-dispersion loss and intra-dispersion loss. 
Specifically, we apply the inter-dispersion loss to the text prompts to spread text embeddings. 
On the other hand, intra-dispersion loss is applied to the visual prompts to minimize the variability of image embeddings of the same class. 

To verify the effectiveness of DAPT, we conducted experiments on few-shot learning and domain generalization tasks with various benchmark datasets. 
For few-shot learning with one sample (1-shot) to 16 samples (16-shots) per class, the proposed method is evaluated on 11 benchmark datasets.   
For domain generalization, 4 benchmark datasets were used after few-shot learning on ImageNet~\cite{deng2009imagenet}. 
Overall, we achieve a significant improvement over recent baselines for few-shot learning and domain generalization.

In summary, we propose DAPT, a prompt tuning method that is aware of the data distributions to improve the performance of VLMs in the few-shot learning setup. 
Unlike the orthodox prompt tuning method, DAPT optimizes text and visual prompts to find the appropriate distribution in each modality. 
In Section~\ref{s:method}, we discuss the details of DAPT and show the various experiments in Section~\ref{s:exp}.
\section{Related Work}
\noindent\textbf{Pre-Trained Vision-Language Models. }
Pre-trained vision-language models (VLMs)~\cite{jia2021scaling, radford2021learning, yuan2021florence, zhang2022contrastive} jointly learn text and image embeddings with large-scale noisy image-text paired datasets. 
Out of those, CLIP~\cite{radford2021learning} and ALIGN~\cite{jia2021scaling} optimize cross-modal representations between positive pairs via contrastive learning and demonstrate impressive performance in various downstream tasks~\cite{Du_2022_CVPR,gu2022openvocabulary,li2023blip2,mokady2021clipcap,zhang2023llamaadapter, zhou2022maskclip}. 
In addition, there are approaches to enhance the ability of VLMs by adjusting latent space in succeeding research. 
For instance, Wang \etal\cite{pmlr-v119-wang20k} claim that alignment and uniformity are two key properties to optimize. 
By expanding these properties, Goel \etal\cite{NEURIPS2022_2cd36d32} propose CyCLIP to mitigate inconsistent prediction in CLIP, fixing the CLIP embedding geometry.

\noindent\textbf{Prompt Tuning. }
Prompting has been studied in natural language processing (NLP). 
Prompt tuning methods such as Petroni \etal~\cite{petroni2019language}, Shin \etal~\cite{shin2020autoprompt}, and Jiang \etal~\cite{jiang2020can} are proposed to construct suitable prompt template. 
Under the influence of NLP, prompt tuning methods with vision-language models are actively studied in the computer vision domain. 
Unlike the hard prompts suggested in CLIP~\cite{radford2021learning}, several works have studied soft prompts by optimizing learnable vectors in text or visual modality. 
CoOp~\cite{zhou2022coop} composes prompt concatenated with label embedding and learnable vectors by the text encoder. 
CoCoOp~\cite{zhou2022conditional} is an advanced version of CoOp and improves generalizability in unseen classes. 
Also, VPT~\cite{jia2022visual} and VP~\cite{bahng2022visual} propose prompt tuning on the visual modality. 
VPT uses learnable vectors for prompt tuning in the Vision Transformer~\cite{dosovitskiy2020vit}. 
Different from prior works, VP suggests image pixel-level prompt tuning in CLIP image encoder. 
Those prompt tuning methods show remarkable transferability and generalizability with only a few parameters. 
More recently, ProDA~\cite{lu2022prompt} and PLOT~\cite{chen2023plot} use multiple prompts and demonstrate better performance than a single text prompt. 
Based on recent success in prompt tuning, there are multimodal prompt tuning methods in VLMs. 
UPT~\cite{zang2022unified} jointly optimize modality-agnostic prompts with extra layers. 
MVLPT~\cite{shen2022multitask} focuses on multi-task prompting. 
MaPLe~\cite{khattakMaPLe} improves generalizability of VLMs via multimodal prompting. 
\section{Method}
\label{s:method}
In this section, we briefly revisit the CLIP~\cite{radford2021learning} and the several prompt tuning methods~\cite{jia2022visual,zhou2022coop} in Section~\ref{s:prelim}. 
Then we propose a distribution-aware prompt tuning, DAPT, in Section~\ref{s:dapt} in detail.

\subsection{Preliminaries}
\label{s:prelim}
CLIP~\cite{radford2021learning} is a vision-language model which trained via contrastive learning on a massive number of image-text pairs. 
In general, CLIP consists of image encoder $f$ and text encoder $g$. 
Given an image $\boldsymbol{x}$ and text label $\boldsymbol{t}$, image embedding $\boldsymbol{z}$ and text embedding $\boldsymbol{w}$ can be obtained as follows: 
\begin{align}
    \label{eq:image_encoder}
    & \boldsymbol{z} = f(\boldsymbol{x}) \\
    \label{eq:text_encoder}
    & \boldsymbol{w} = g(\boldsymbol{t}). 
\end{align}
Note that image embedding $\boldsymbol{z}$ and text embedding $\boldsymbol{w}$ are normalized.  
Given $C$ image classes, the prediction probability can be calculated by softmax with the cosine similarity between the image embeddings and the corresponding text embeddings representing the image class given as:
\begin{equation}
\label{eq:cosine_sim}
\begin{split}
    & p(y=c|\boldsymbol{x}) = \cfrac{\exp{(\boldsymbol{z}^\top\boldsymbol{w}_c/\tau})}{{\sum_{j=1}^C}\exp{(\boldsymbol{z}^\top \boldsymbol{w}_j/\tau})},
\end{split}
\end{equation}
where $\tau$ is a temperature parameter, and 
$\boldsymbol{w}_c$ represents the text embedding of the class label $\boldsymbol{t}_c$. 
Combining with cross-entropy, we define CLIP loss, $\mathcal{L}_{\text{CLIP}}$, as follows:
\begin{equation}
\label{eq:clip_loss}
\begin{split}
    & \mathcal{L}_{\text{CLIP}} = -\frac{1}{B}\sum_{i=1}^B\log\frac{\exp{(\boldsymbol{z}_i^\top\boldsymbol{w}_{y_i}/\tau)}}{\sum_{j=1}^C{\exp{(\boldsymbol{z}_i^\top\boldsymbol{w}_{j}/\tau}})},
\end{split}
\end{equation}
where $y_i$ denotes the class of $i$-th image $\boldsymbol{x}$, and $B$ is the batch of image-text pairs. 

\noindent\textbf{Text Prompt.} 
CoOp~\cite{zhou2022coop} is the first approach to apply prompt tuning in the text encoder of CLIP. 
In CoOp, text prompt $\boldsymbol{p}$ is represented as a learnable vector $v$ combined with the class. 
Then, the input of the text encoder is given as: 
\begin{align}
    \label{eq:text_prompt}
    & \boldsymbol{p}_j = [v_1, v_2, \cdots, v_L,\texttt{CLASS}].
\end{align}
The output of the text encoder with soft prompts is represented as:
\begin{align}
    \label{eq:text_feature}
    & \tilde{\boldsymbol{w}}_j = g(\boldsymbol{p}_j). 
\end{align}
Note that ${\tilde{\boldsymbol{w}}}$ is normalized. 
CoOp uses various configurations with respect to the length $L$ and positions depending on datasets. 
They can be viewed as hyperparameters for prompt tuning. 
In our method, we fixed the hyperparameters for all settings. The learnable vectors of the text prompt are placed in front of $\texttt{CLASS}$ with a length of $L = 16$. 

\noindent\textbf{Visual Prompt.} 
In the computer vision domain, VPT~\cite{jia2022visual} proposed a visual prompt tuning method for Vision Transformers (ViT)~\cite{dosovitskiy2020vit}. 
Similar to CoOp, VPT inserts the learnable vector $u$ between class token $\mathtt{CLS}$ and image patch embeddings $\boldsymbol{E}$ for the image encoder. 
Since CLIP~\cite{radford2021learning} uses ViT backbone for the image encoder, we define the visual prompt in CLIP as below: 
\begin{equation}
\begin{split}
    \label{eq:visual_prompt}
    & \boldsymbol{q}_i = [\mathtt{CLS}, u_1, u_2 \cdots, u_L, \boldsymbol{E}].
\end{split}
\end{equation}
We set the length of learnable vectors of the visual prompt to $L = 16$, which is the same as the text prompt in~\eqref{eq:text_prompt}. 
From~\eqref{eq:visual_prompt}, we can obtain output image embedding $\tilde{\boldsymbol{z}}_i$ with visual prompt $\boldsymbol{q}_i$ as:
\begin{equation}
\begin{split}
    \label{eq:visual_feature}
    & \tilde{\boldsymbol{z}}_i = f(\boldsymbol{q}_i).
\end{split}
\end{equation}
Note that ${\tilde{\boldsymbol{z}}}$ is normalized.

\noindent\textbf{Prompt Tuning.} 
When fine-tuning CLIP with prompts, the image encoder and the text encoder have typically frozen the weights of all layers.
Therefore, only the prompts are optimized. 
For large-scale pre-trained models, prompt tuning is often more effective and efficient than traditional fine-tuning methods such as linear probing and full fine-tuning of all layers. 

\subsection{Distribution-Aware Prompt Tuning}
\label{s:dapt}
\begin{figure*}[tb]
    \centering
    \begin{tabular}{cc}
    \multirow{3}{*}[6em]{\subfloat[The architecture of DAPT]{\includegraphics[width=0.4\textwidth]{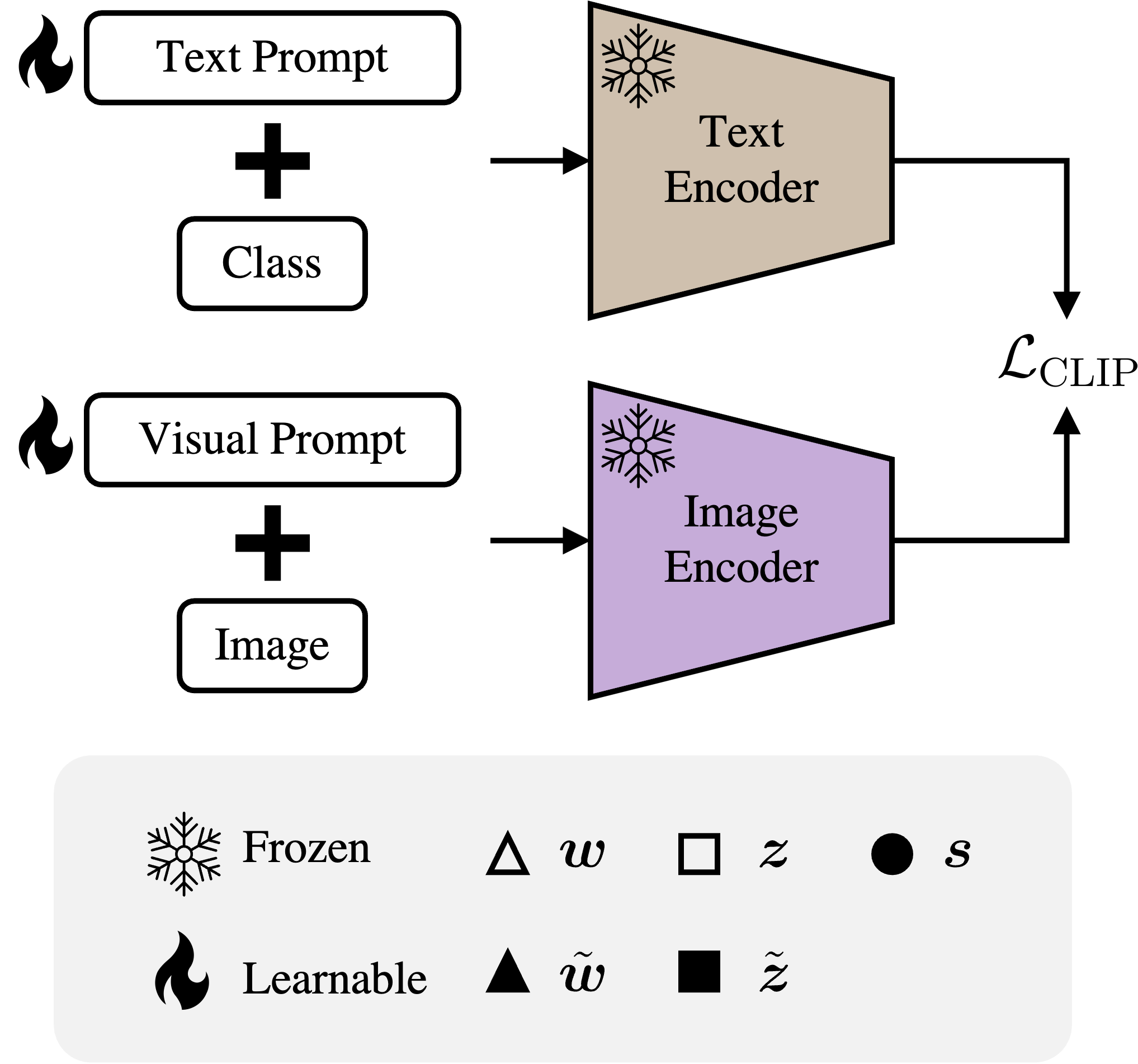}
    \label{fig:arch_arch}}} 
    & \subfloat[Inter-dispersion with text prompt]{\includegraphics[width=0.48\textwidth]{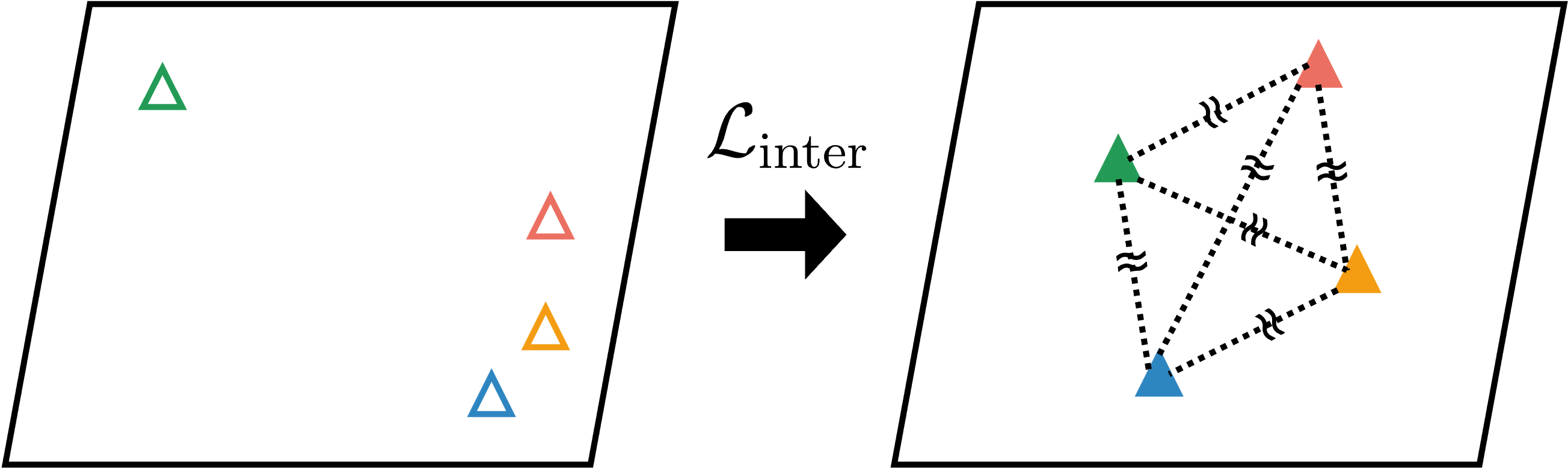}
    \label{fig:arch_inter} 
    } \\ & \\
    & \subfloat[Intra-dispersion with visual prompt]{\includegraphics[width=0.48\textwidth]{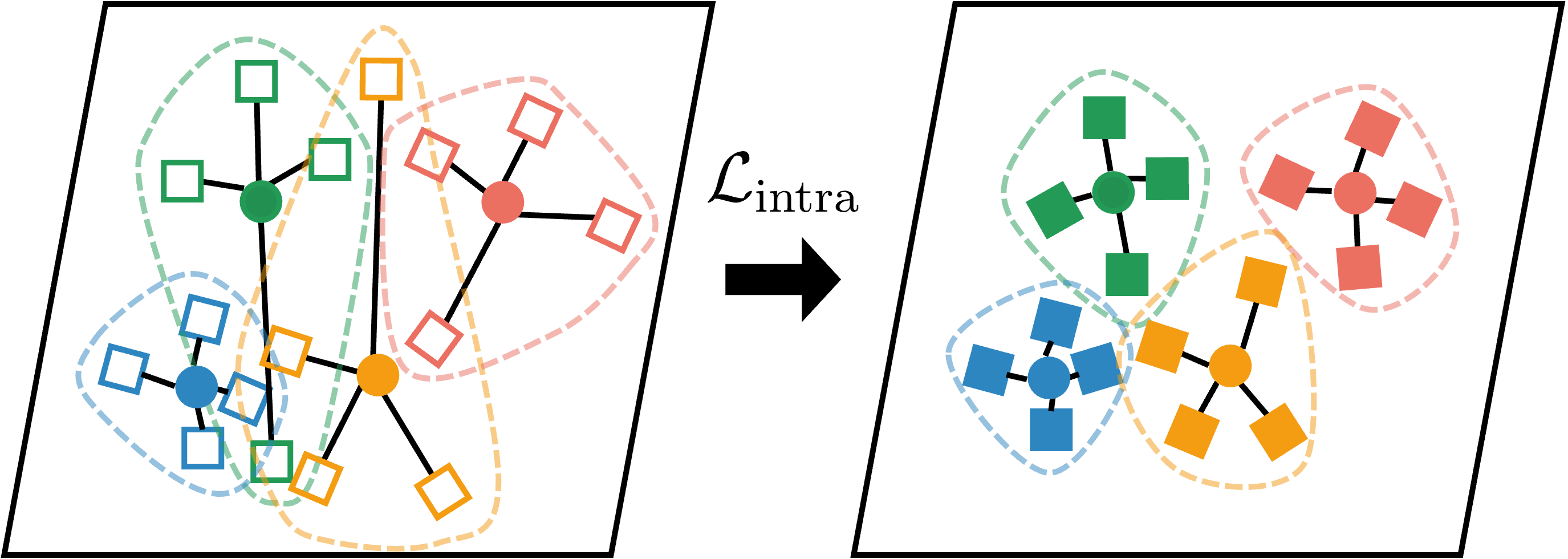}
    \label{fig:arch_intra} } \\ 
\end{tabular}
\caption{\textbf{Overall architecture of DAPT}. 
(a) DAPT consists of CLIP~\cite{radford2021learning} architecture combined with CoOp~\cite{zhou2022coop} and VPT~\cite{jia2022visual}. The symbols mean text and visual output embedding (\ie, $\boldsymbol{w}$ and $\boldsymbol{z}$), text and visual outputs combined with prompts (\ie, $\tilde{\boldsymbol{w}}$ and $\tilde{\boldsymbol{z}}$), and the prototype $\boldsymbol{s}$. Following prompt tuning manner, text and image encoders are frozen during training, and only prompts are updated. 
(b) Inter-dispersion loss $\mathcal{L}_\text{inter}$ defined from Gaussian potential kernel $\mathrm{G}$ is applied to text prompts to expand the distance between each text embeddings $\tilde{\boldsymbol{w}}$ to avoid embedding collapse. 
(c) To aggregate image embeddings within the same class, we define the prototype $\boldsymbol{s}$ demonstrates representative image embeddings of each class by calculating the average of zero-shot CLIP image embeddings $\boldsymbol{z}$.
Then, intra-dispersion loss $\mathcal{L}_\text{intra}$ is applied to the visual prompt to gather image embeddings around the prototype $\boldsymbol{s}$. }
\label{fig:architecture}
\end{figure*}
We present DAPT that improves feature alignment between text and visual modalities 
by optimizing the distributions of embeddings via inter-dispersion and intra-dispersion losses.
The overall pipeline of the proposed method is presented in Figure~\ref{fig:architecture} and how inter-dispersion and intra-dispersion losses optimize text and visual latent spaces is depicted in Figure~\ref{fig:arch_inter} and~\ref{fig:arch_intra}, respectively.

\noindent\textbf{Inter-Dispersion Loss for Text Prompts.}
A small distance between text (label) embeddings  may lead to misclassification and make it difficult to align visual features.
To address this issue, we introduce an inter-dispersion loss in text embeddings based on the \textit{uniformity} inspired by Wang~\etal~\cite{pmlr-v119-wang20k}. \textit{Uniformity} means that the feature embeddings are roughly uniformly distributed in the hypersphere. 
This minimizes the overlap between embeddings and enables better alignment.
With normalized text embeddings $\tilde{\boldsymbol{w}}$, we define Gaussian potential kernel $\mathrm{G}$ as follows: 
\begin{equation}
\begin{split}
    \label{eq:rbf}
    & \mathrm{G}(\tilde{\boldsymbol{w}}_m, \tilde{\boldsymbol{w}}_n):=\exp{(-t\Vert\tilde{\boldsymbol{w}}_m-\tilde{\boldsymbol{w}}_n\Vert^2_2)},
\end{split}
\end{equation}
where $m, n \in C$ and $m \neq n$.

Minimizing the Gaussian potential kernel $\mathrm{G}$ above increases the distance between the text embeddings of prompt $\boldsymbol{p}_m$ and $\boldsymbol{p}_n$ on the hypersphere.
To optimize the distribution of text embeddings encouraging \textit{uniformity}, we define the inter-dispersion loss as follows: 
\begin{align}
    \mathcal{L}_\text{inter}&=\sum_{m \neq n}\mathrm{G}(\tilde{\boldsymbol{w}}_m, \tilde{\boldsymbol{w}}_n)\\
    \label{eq:inter_loss}
    &=\sum_{m \neq n}\exp{(-t{\Vert\tilde{\boldsymbol{w}}_m - \tilde{\boldsymbol{w}}_n \Vert}_2^2)}.
\end{align}
Note that we set hyperparameter $t=2$ for all experiments in this paper. 

\noindent\textbf{Intra-Dispersion Loss for Visual Prompts.}
Given a class, unlike the uniquely defined text (label) embedding, multiple visual embeddings exist in the latent space.
Specifically, due to multiple images per class in the dataset, various image embeddings are obtained from the image encoder.

For better alignment between the text and image embeddings given class $\boldsymbol{t}_c$, image embeddings of the same class should be close to each other.
To reduce the intra-class distance of image embeddings $\tilde{\boldsymbol{z}}_i$ and $\tilde{\boldsymbol{z}}_j$, 
we define the prototype $\boldsymbol{s}$ motivated by PROTONET~\cite{NIPS2017_cb8da676} with training samples $\mathcal{D}_N = \{(\boldsymbol{x}_i, \boldsymbol{y}_i)\}_{i=1}^N$ as follows: 
\begin{equation}
\begin{split}
    \label{eq:prototype}
    & \boldsymbol{s}_c = \cfrac{1}{N}\sum_{(\boldsymbol{x}_i, \boldsymbol{y}_i)\in\mathcal{D}_N^c}{\boldsymbol{z}_i},
\end{split}
\end{equation}
where $\boldsymbol{z}_i = f(\boldsymbol{x}_i)$. 
Note that $\mathcal{D}^c_{N} = \{(\boldsymbol{x}_i, \boldsymbol{y}_i) \in \mathcal{D}_{N} \vert \boldsymbol{y}_i=c\}$ and $N$ is the number of training samples. 
In order to cluster image embeddings with the same class, we assume that each embedding should be close to its prototype. 
Therefore, the intra-dispersion loss $\mathcal{L}_\text{intra}$, which reduces the distance between the image embedding and the prototype $\boldsymbol{s}$, is defined as follows: 
\begin{equation}
\begin{split}
    \label{eq:intra_loss}
    & \mathcal{L}_\text{intra} = \sum_c \sum_i \mathbbm{1}_{[y_i = c]} {\Vert \tilde{\boldsymbol{z}}_i - \boldsymbol{s}_c \Vert}_2^2 ,
\end{split}
\end{equation}
where $c$ is the corresponding class index of input image $\boldsymbol{x}_i$.

\noindent\textbf{Optimization.}
Combining the CLIP loss in~\eqref{eq:clip_loss},  our dispersion losses in~\eqref{eq:inter_loss} and~\eqref{eq:intra_loss}, DAPT optimizes text prompt $\boldsymbol{p}$ from ~\eqref{eq:text_prompt} and visual prompt $\boldsymbol{q}$ from ~\eqref{eq:visual_prompt} by minimizing the following total loss: 
\begin{equation}
\begin{split}
    \label{eq:total_loss}
    & \mathcal{L} =\mathcal{L}_\text{CLIP}+\beta_t\mathcal{L}_\text{inter}+\beta_v\mathcal{L}_\text{intra} ,
\end{split}
\end{equation}
where $\beta_t$ and $\beta_v$ are hyperparameters for each dispersion loss. 
\begin{algorithm}[th]
\footnotesize
\caption{DAPT}
\label{algo:dapt}
\begin{algorithmic}[1]
\Require Pre-trained CLIP image encoder $f$ and text encoder $g$, dataset $\mathcal{D}_N$ with $C$ classes
\State $\boldsymbol{z}_i \gets~f(\boldsymbol{x_i})$
\Comment{See~\eqref{eq:image_encoder}.}
\State $\boldsymbol{s}_c \gets~\cfrac{1}{N}\sum\limits_{(\boldsymbol{x}_i, \boldsymbol{y}_i)\in\mathcal{D}_N^c}{\boldsymbol{z}_i}, \text{for} \;\forall c$ \Comment{See~\eqref{eq:prototype}.}
\State $\tilde{\boldsymbol{z}}_i \gets~ f(\boldsymbol{q}_i)$ \Comment{See ~\eqref{eq:visual_prompt} and \eqref{eq:visual_feature}.}
\State $\tilde{\boldsymbol{w}}_j \gets~ g(\boldsymbol{p}_j)$ \Comment{See~\eqref{eq:text_prompt} and~\eqref{eq:text_feature}.}
\For{$\mathcal{D}(\boldsymbol{x}_i, \boldsymbol{y}_i)$}
\State $\mathcal{L}_\text{CLIP} \gets~-\cfrac{1}{B}\sum\limits_{i=1}^B\log\cfrac{\exp{(\tilde{\boldsymbol{z}}_i^\top \tilde{\boldsymbol{w}}_{y_i}/\tau})}{\sum_{j=1}^C{\exp{(\tilde{\boldsymbol{z}}_i^\top \tilde{\boldsymbol{w}}_{j}/\tau}})}$ \Comment{See~\eqref{eq:clip_loss}.}
\State $\mathcal{L}_\text{inter} \gets~ \sum\limits_{m \neq n}\exp{(-t{\Vert \tilde{\boldsymbol{w}}_m - \tilde{\boldsymbol{w}}_n \Vert}_2^2)}$ \Comment{See~\eqref{eq:inter_loss}.}
\State $\mathcal{L}_\text{intra} \gets~ \sum\limits_c \sum\limits_i \mathbbm{1}_{[y_i = c]} {\Vert \tilde{\boldsymbol{z}}_i - \boldsymbol{s}_c \Vert}_2^2$ \Comment{See ~\eqref{eq:intra_loss}.}
\State $\mathcal{L} \gets~ \mathcal{L}_\text{CLIP} + \beta_t\mathcal{L}_\text{inter} + \beta_v\mathcal{L}_\text{intra}$
\State $\mathcal{L}$.\texttt{backward()}
\EndFor
\State $\tilde{\boldsymbol{z}}$.\texttt{update()}
\State $\tilde{\boldsymbol{w}}$.\texttt{update()}
\end{algorithmic}
\end{algorithm}
\\

Algorithm~\ref{algo:dapt} summarizes how DAPT optimizes text and visual prompts regarding the distribution of each modality in latent spaces by minimizing the proposed loss in~\eqref{eq:total_loss}. 
To sum up, during training, text prompt $\tilde{\boldsymbol{w}}$ and visual prompt $\tilde{\boldsymbol{z}}$ are optimized via combined loss which consists of inter-dispersion loss, intra-dispersion loss, and CLIP loss.
\section{Experiments}
\label{s:exp}
\noindent\textbf{Datasets. } 
We evaluate DAPT on few-shot image classification and domain generalization settings. 
We evaluate 11 public datasets in few-shot learning, 
Food101~\cite{bossard2014food}, 
DTD~\cite{cimpoi2014describing}, 
Imagenet~\cite{deng2009imagenet}, 
Caltech101~\cite{fei2004learning}, 
EuroSAT~\cite{helber2019eurosat}, 
StanfordCars~\cite{krause20133d}, 
FGVCAircraft~\cite{maji13fine-grained}, 
Flowers102~\cite{nilsback2008automated}, 
OxfordPets~\cite{parkhi2012cats}, 
UCF101~\cite{soomro2012ucf101}, 
and SUN397~\cite{xiao2010sun}, using 1, 2, 4, 8, and 16-shots per dataset. 
In the domain generalization setting, we set the source dataset to ImageNet and test to target dataset - ImageNet-R~\cite{hendrycks2021many}, ImageNet-A~\cite{hendrycks2021natural},  ImageNetV2~\cite{recht2019imagenet}, and ImageNet-Sketch~\cite{wang2019learning}. 

\noindent\textbf{Baselines. }
In the experiments, we compare with zero-shot CLIP, linear probe CLIP, CoOp~\cite{zhou2022coop}, and VPT~\cite{jia2022visual}. 
In the case of zero-shot CLIP, we test with pre-trained CLIP without additional training. 
On the other hand, we fine-tune the classifier in the case of linear probe CLIP following Radford \etal~\cite{radford2021learning}. 
Because we demonstrate DAPT on CLIP with the ViT-B/16 image encoder backbone, we implement CoOp and VPT with ViT-B/16. 
Additionally, we observed that VPT has various accuracy gaps in the few-shot learning setting. 
For a fair comparison, we search hyperparameters (\textit{i.e.}, learning rate) with ranges from 0.002 to 20, as reported in Table~\ref{tab:vpt_lr}. 
The figures show the average accuracy of 11 datasets in 16-shots image classification settings for each learning rate. 
As a result, we set the learning rate is 0.2 on VPT in all experiments. 

\begin{table}[h]
\centering
\begin{tabular}{c c c c c c}
\toprule
Learning rate & 0.002 & 0.02 & 0.2 & 2.0 & 20.0 \\
\midrule
VPT & 68.32 & 73.72 & \textbf{76.56} & 76.40 & 76.13 \\
\bottomrule
\end{tabular}
\caption{Hyperparameter search on VPT.}
\label{tab:vpt_lr}
\end{table}
\noindent\textbf{Implementation Details. }
We use pre-trained CLIP~\cite{radford2021learning} with ViT-B/16 image encoder backbone from the official repository\footnote{https://github.com/openai/CLIP}. 
To construct prompts for text and visual encoders, we refer to CoOp and VPT open sources\footnote{https://github.com/KaiyangZhou/CoOp}$^{,}$\footnote{https://github.com/KMnP/vpt}. 
In all experiments, we evaluate three times on NVIDIA RTX-3090 GPUs and report the average value.  
More implementation details are included in the supplement. 

\subsection{Few-Shot Learning}
\begin{figure*}[tb]
    \centering
    \begin{tabular}{cc}
    \subfloat[DAPT vs. CoOp]{\includegraphics[width=0.48\textwidth]{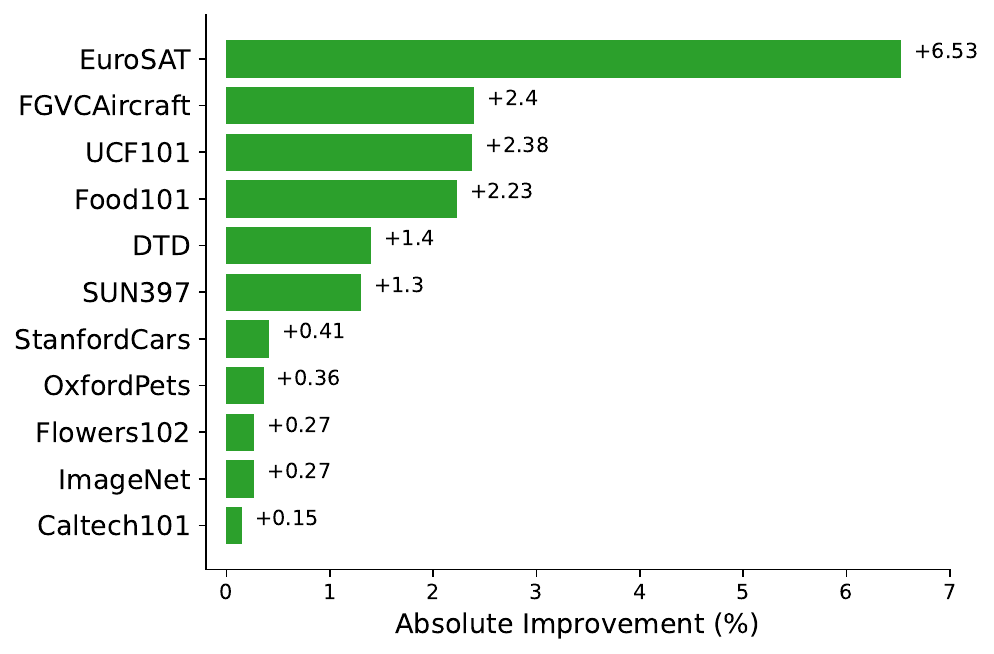} \label{fig:bar_coop}}
     & 
    \subfloat[DAPT vs. VPT]{\includegraphics[width=0.48\textwidth]{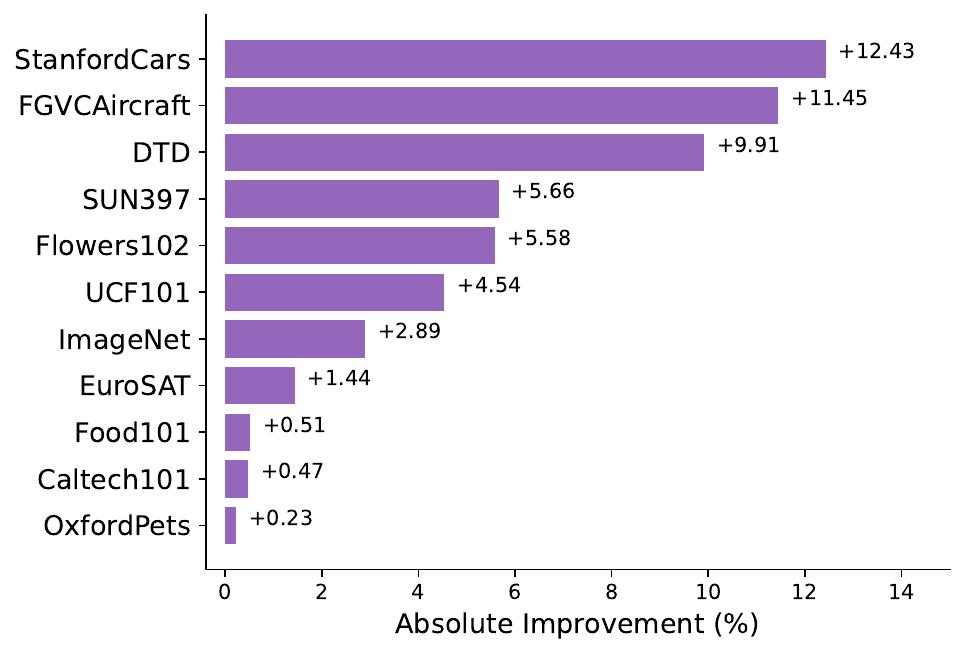} \label{fig:bar_vpt}} \\ 
\end{tabular}
\caption{\textbf{Comparison with CoOp and VPT}. We show the result of image classification with 16 samples from training data per each dataset. Our method has overall improvement compared with CoOp~\cite{zhou2022coop} and VPT~\cite{jia2022visual}.}
\label{fig:bar}
\end{figure*}

Figure~\ref{fig:graph} summarizes the performance of DAPT in few-shot learning on 11 datasets and the average accuracy.
Each plot compares DAPT with baselines. 
The experiments show DAPT outperforms baselines on most benchmark datasets. 
In addition, Figure~\ref{fig:bar} also shows that DAPT consistently outperforms revious prompt tuning methods - CoOp~\cite{zhou2022coop} and VPT~\cite{jia2022visual} on all datasets. 

\begin{table}[ht]
\centering
\begin{tabular}{lrrr}
\toprule
\multirow{1}{*}{Dataset}& \multicolumn{1}{c}{DAPT} & \multicolumn{1}{c}{Linear probe CLIP} & \multicolumn{1}{c}{$\Delta$}\\ 
\midrule
OxfordPets & 89.55 & 43.33 & 46.22 \\
Flowers102 & 76.83 & 72.11 & 4.72 \\
FGVCAircraft & 4.44 & 19.62 & -15.18 \\
DTD & 20.71 & 34.89 & -14.18 \\
EuroSAT & 38.54 & 49.88 & -9.88 \\
StanfordCars & 67.39 & 36.04 & 31.35 \\
Food101 & 82.27 & 45.99 & 36.28 \\
SUN397 & 67.10 & 41.87 & 25.23 \\
Caltech101 & 92.20 & 81.12 & 11.08 \\
UCF101 & 71.71 & 54.14 & 17.57 \\
ImageNet & 65.00 & 32.99 & 32.01 \\ 
Average & 61.42 & 46.54 & 14.87 \\ 
\bottomrule
\end{tabular}
\caption{
\textbf{Comparison with Linear probe CLIP.}  We report image classification with 1-shot. $\Delta$ represents difference between DAPT and Linear probe CLIP. }
\label{tab:lpclip_1shot}
\end{table}
\begin{figure}[h!]
\centering
\includegraphics[width=0.95\linewidth]{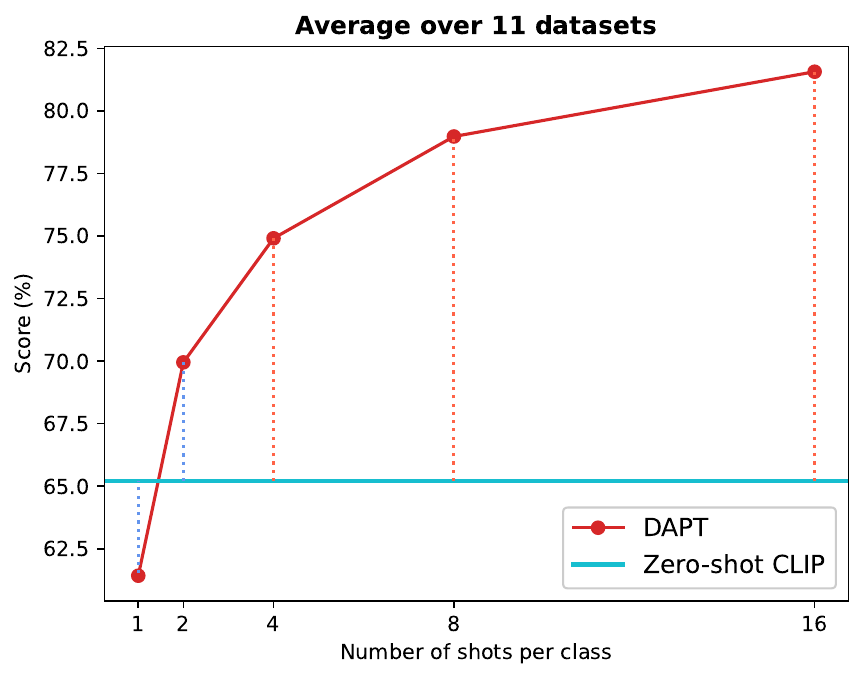}
\caption{Comparison with Zero-shot CLIP.}
\label{fig:zsclip}
\end{figure}
\noindent\textbf{Comparison with Linear Probe CLIP. }
We observed that DAPT shows higher performance than linear probe CLIP for most datasets as noted in Table~\ref{tab:lpclip_1shot}. 
Especially on OxfordPets~\cite{parkhi2012cats}, DAPT improves the performance by 46.22\% with only a single training sample. 
When comparing all experiments in 1-shot image classification, DAPT achieves an average improvement of 14.87\% compared to linear probe CLIP. 

\noindent\textbf{Comparison with Zero-Shot CLIP. }
When comparing DAPT to zero-shot CLIP, DAPT demonstrates higher performance in most experiments, with a similar result to the comparison with linear probe CLIP. 
Notably, as described in Figure~\ref{fig:zsclip}, the performance gap between DAPT and zero-shot CLIP widens as the number of samples increases, suggesting that DAPT is adaptive at distribution-aware prompt optimization as the training sample size grows.

\begin{figure*}[tbh]
\centering
\begin{tabular}{r c l}
    \includegraphics[width=.31\linewidth]{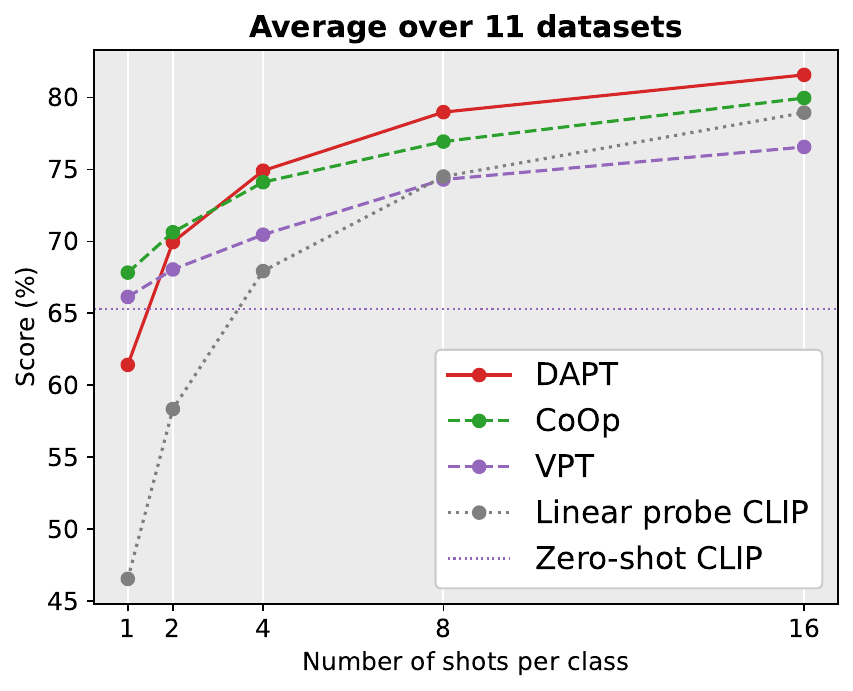} & \includegraphics[width=.31\linewidth]{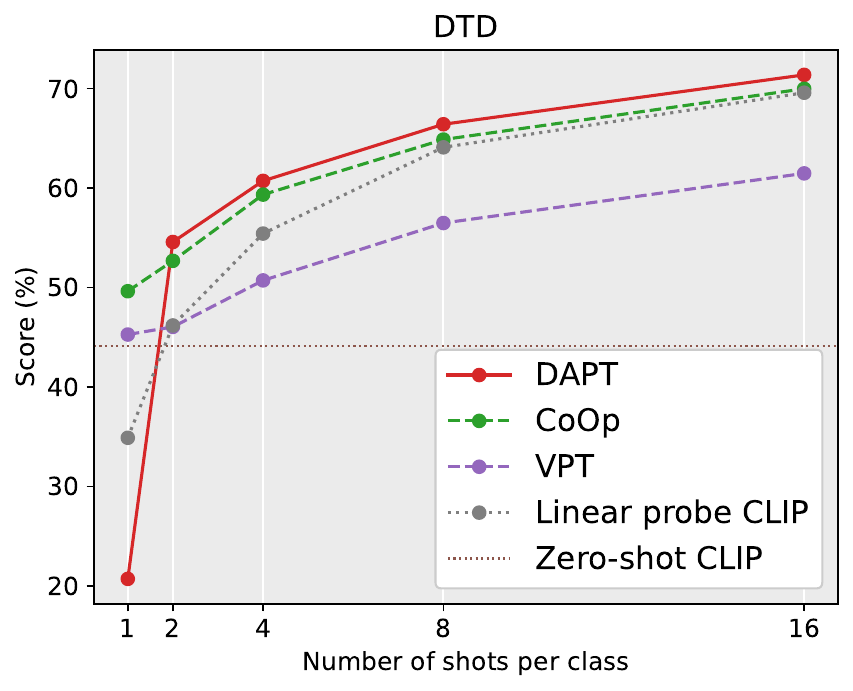} & \includegraphics[width=.31\linewidth]{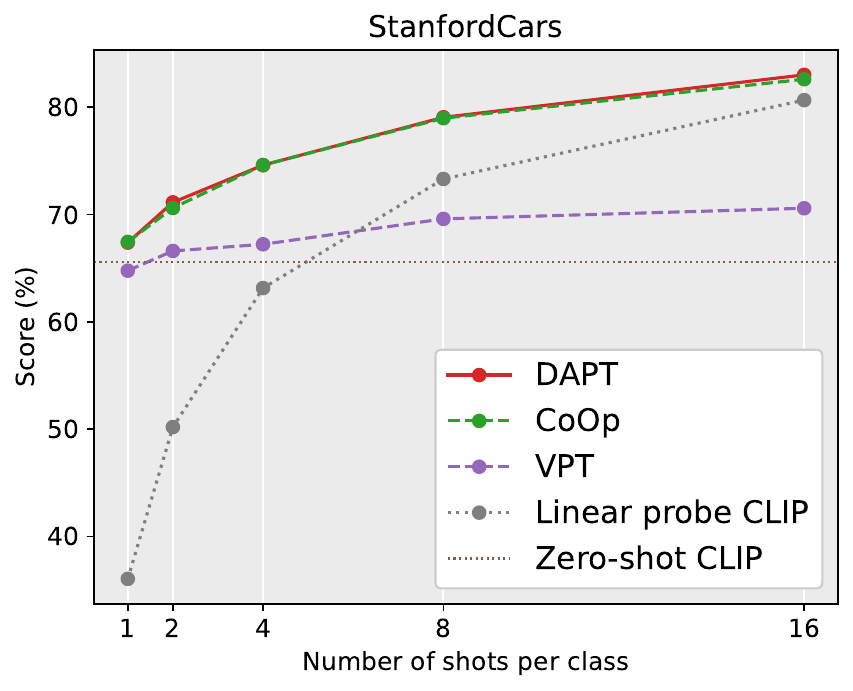} \\ 
    \includegraphics[width=.31\linewidth]{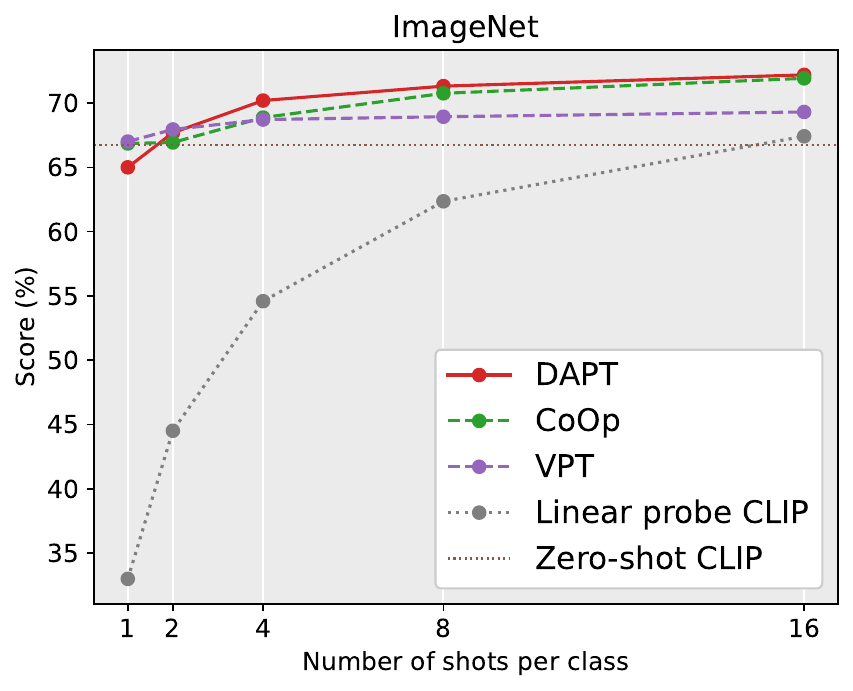} & \includegraphics[width=.31\linewidth]{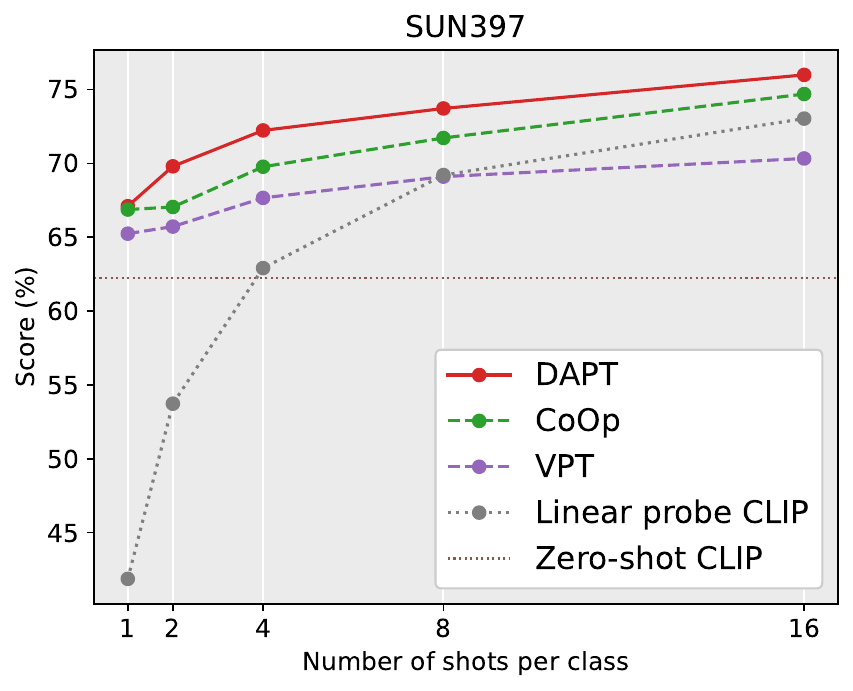} & 
    \includegraphics[width=.31\linewidth]{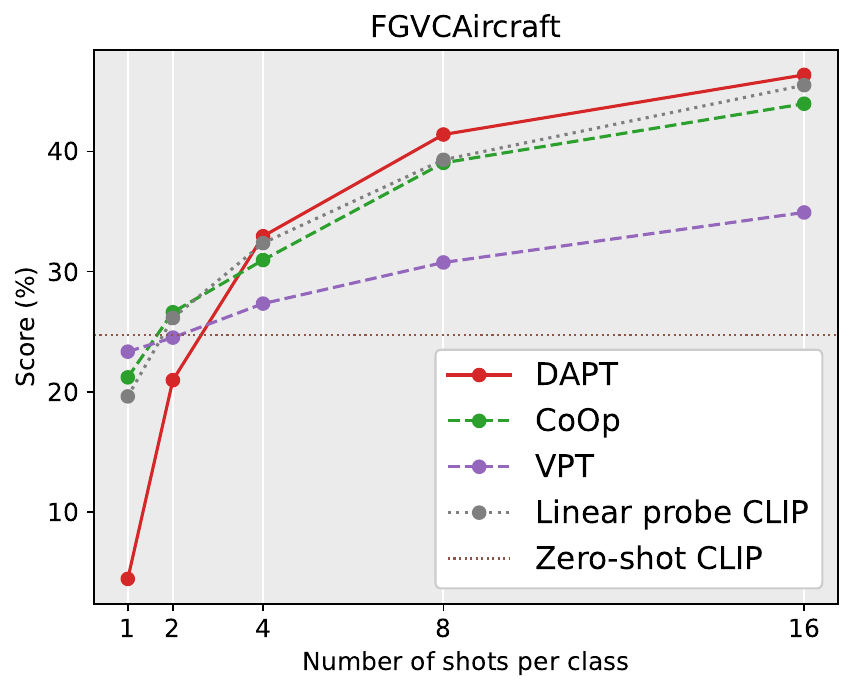} \\ 
    \includegraphics[width=.31\linewidth]{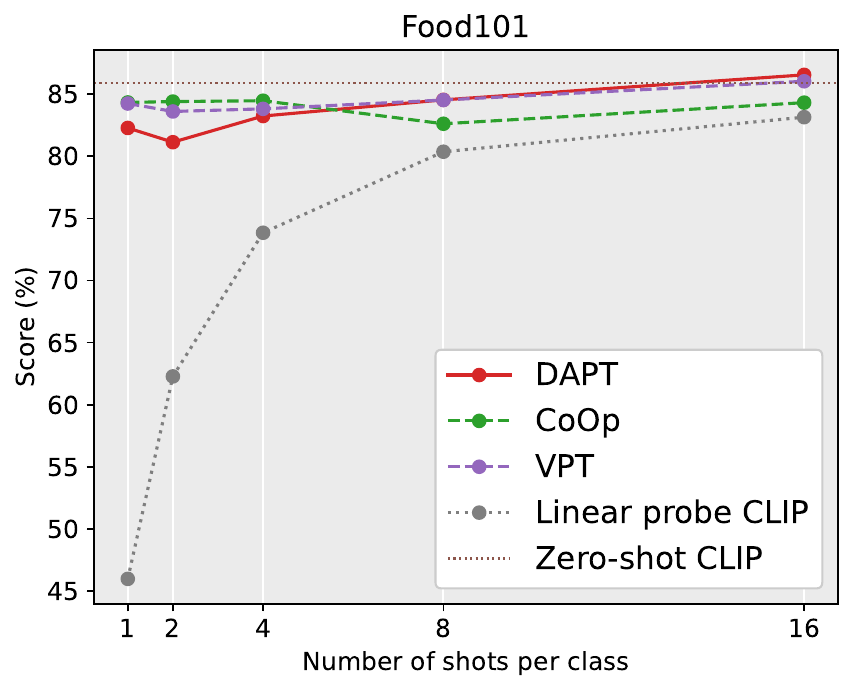} & \includegraphics[width=.31\linewidth]{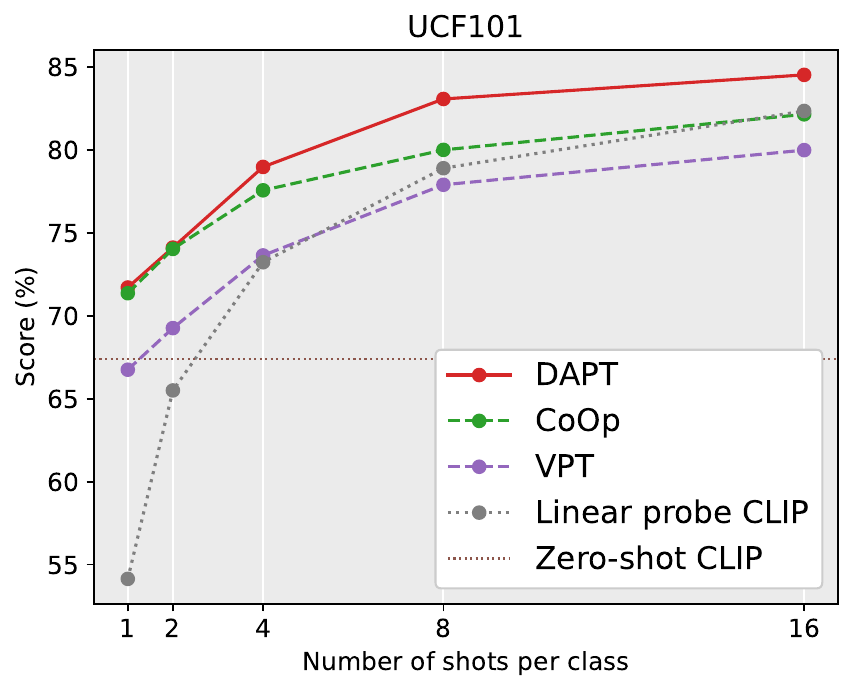} & \includegraphics[width=.31\linewidth]{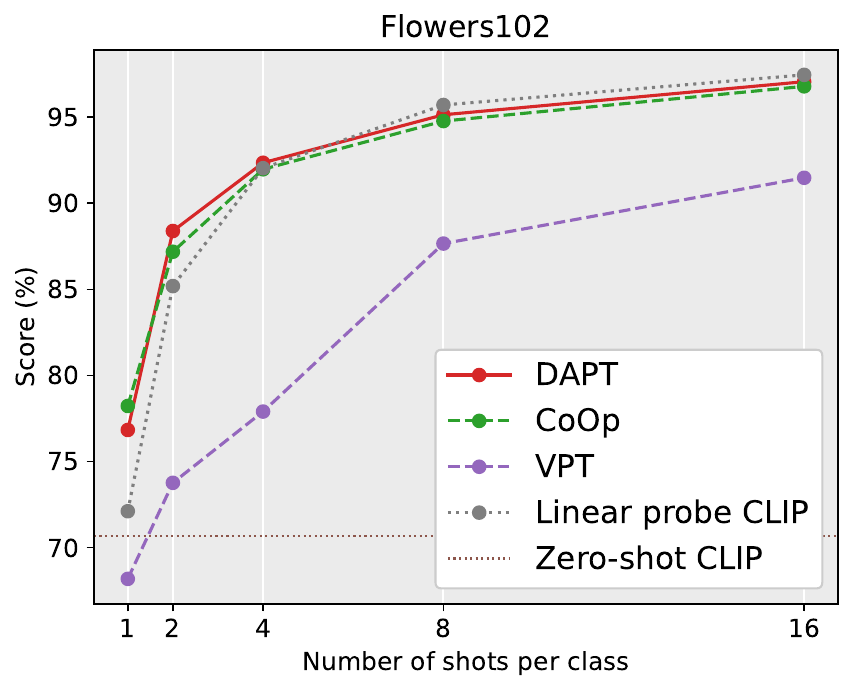} \\ 
    \includegraphics[width=.31\linewidth]{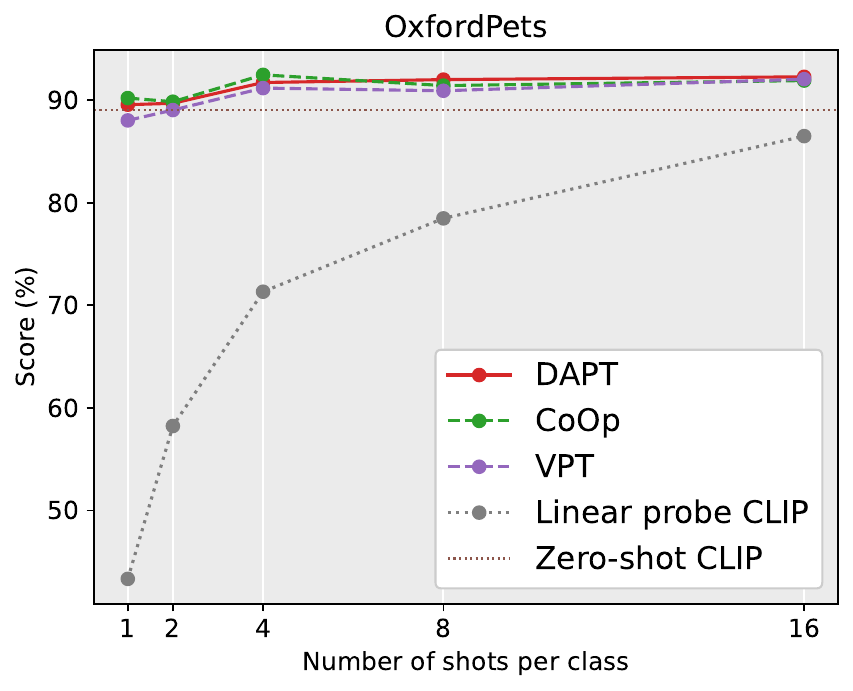} & \includegraphics[width=.31\linewidth]{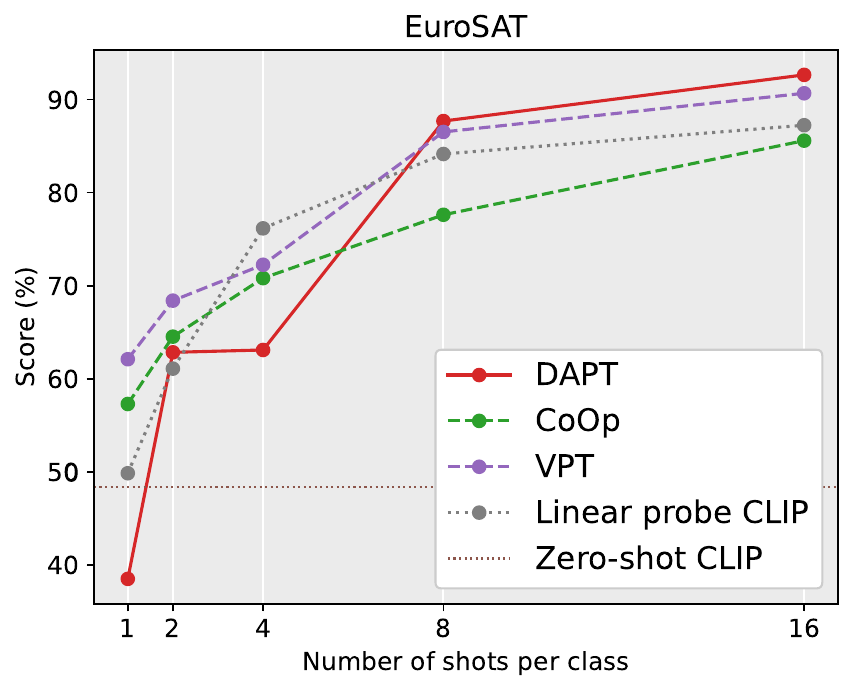} & \includegraphics[width=.31\linewidth]{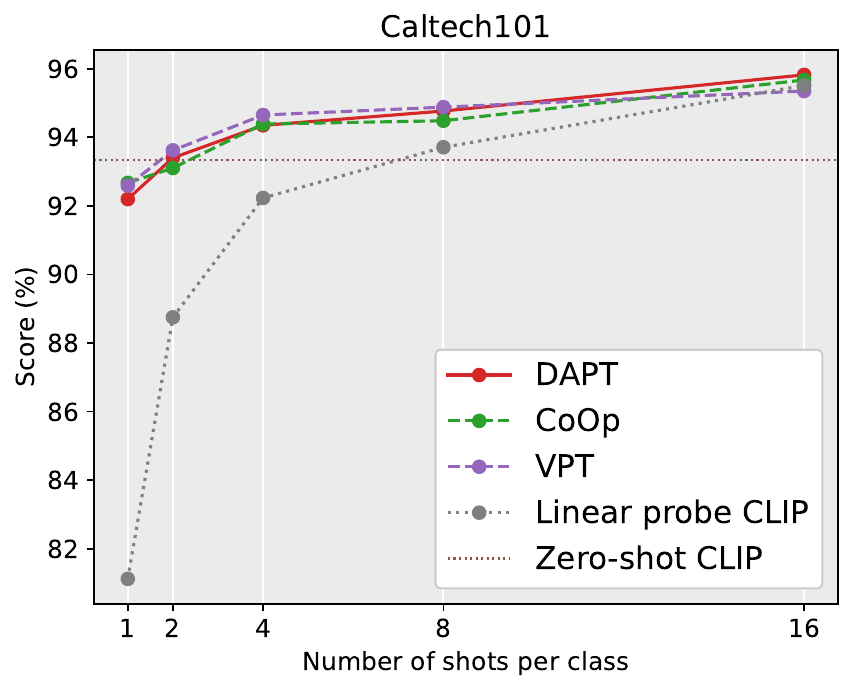} \\ 
\end{tabular}
\caption{Few-shot learning in image classifications on 11 datasets.}
\label{fig:graph}
\end{figure*}

\noindent\textbf{Comparison with Previous Prompt Tuning. }
Figure~\ref{fig:bar_coop} demonstrates the results of DAPT and CoOp~\cite{zhou2022coop} on 16 shots. 
As shown in the results, DAPT outperforms CoOp for all datasets. Furthermore, Figure~\ref{fig:bar_vpt} presents the comparison of DAPT and VPT~\cite{jia2022visual} under the same experimental setting. 
As with the CoOp comparison results, it shows that DAPT outperforms VPT on all datasets, and the accuracy is 12.43\% higher than VPT on StanfordCars~\cite{krause20133d}.
To sum up, using DAPT for each modality shows superior performance compared to conventional prompt tuning, as well as zero-shot CLIP and liner probe CLIP.

\subsection{Domain Generalization}
We evaluate the generalizability of DAPT by comparing it with zero-shot CLIP, and linear probe CLIP in the domain generalization setting. 
We use ImageNet~\cite{deng2009imagenet} as a source dataset,and the prompts are trained with 16 samples and the target dataset as ImageNet-R~\cite{hendrycks2021many}, ImageNet-A~\cite{hendrycks2021natural}, ImageNetV2~\cite{recht2019imagenet}, and ImageNet-Sketch~\cite{wang2019learning}. 
\begin{table}[tbh]
\centering
\resizebox{\columnwidth}{!}{
\begin{tabular}{l c c c c c}
\toprule
\multirow{2}{*}[-0.1em]{Method} & Source & \multicolumn{4}{c}{Target} \\
\cmidrule{2-2} \cmidrule(l){3-6}
& ImageNet & -V2 & -Sketch & -A & -R \\
\midrule
Zero-shot CLIP & 66.72 & 60.90	& 46.10 & 47.75 & 73.97 \\
Linear probe CLIP & 67.42 & 57.19 & 35.97 & 36.19 & 60.10 \\
CoOp & 71.93 & 64.22 & 47.07 & \textbf{48.97} & 74.32 \\
VPT & 69.31 & 62.36 & 47.72 & 46.20 & \textbf{75.81} \\
DAPT & \textbf{72.20} & \textbf{64.93} & \textbf{48.30} & 48.74 & 75.75 \\
\bottomrule
\end{tabular}}
\caption{Comparison on domain generalization.}
\label{tab:imagenet_generalization}
\end{table}
\noindent The overall results are shown in Table~\ref{tab:imagenet_generalization}. 
From the experimental results, DAPT achieves remarkable performance on unseen data compared to zero-shot CLIP and linear probe CLIP. 
Compared with CoOp~\cite{zhou2022coop} and VPT~\cite{jia2022visual}, it slightly decreases accuracy on ImageNet-A and ImageNet-R, respectively. In contrast, in the rest of the dataset, ImageNet-V2 and ImageNet-Sketch, DAPT has superior performance with a significant accuracy gain.

\subsection{Ablation Study}
\noindent\textbf{Effectiveness of Intra-dispersion Loss and Inter-dispersion Loss. } 
To verify the accuracy improvement when applying the inter-dispersion loss and intra-dispersion loss, we test few-shot learning experiments on 11 datasets with 16 samples from training data. 
We set the model combined with text prompts, \textit{i.e.}, CoOp~\cite{zhou2022coop}, and visual prompts, \textit{i.e.}, VPT~\cite{jia2022visual}, as a baseline. 
Table~\ref{tab:ablation_loss} shows the performance improvement of applying inter-dispersion loss and intra-dispersion loss. 
The experimental results indicate that the accuracy improved overall for 11 datasets and the average of the entire datasets compared to the baseline. 
As a result, both inter-dispersion loss and intra-dispersion loss show performance gain by reconstructing the embedding space across most datasets. 
In particular, the inter-dispersion loss and intra-dispersion loss are effective for improving performance on FGVCAircraft~\cite{maji13fine-grained}, and DTD~\cite{cimpoi2014describing}, respectively. 
Interestingly, despite some datasets, \textit{i.e.}, Food101~\cite{bossard2014food}, ImageNet~\cite{deng2009imagenet}, FGVCAircraft~\cite{maji13fine-grained}, Flowers102~\cite{nilsback2008automated}, and SUN397~\cite{xiao2010sun}, slightly decreasing accuracy by adding inter-dispersion loss or intra-dispersion loss, DAPT increases accuracy. 
\begin{table}[h]
\centering
\begin{tabular}{l c c c c c}
    \toprule
    \multirow{2}{*}[-0.2em]{Method} & \multicolumn{5}{c}{\# of training samples in each class} \\
    \cmidrule(l){2-6}
    & 1 & 2 & 4 & 8 & 16 \\
    \midrule
    CoOp+VPT & 61.05 & 68.49 & 73.28 & 78.76 & 81.25 \\
    DAPT & \textbf{61.42} & \textbf{69.95} & \textbf{74.91} & \textbf{78.98} & \textbf{81.62} \\
    \bottomrule
\end{tabular}
\caption{\textbf{Ablation study.} The average accuracy of CoOp+VPT and DAPT on 11 benchmark datasets is presented. 
}
\label{tab:ablation_shots}
\end{table}
The results support that reconstructing and optimizing embedding jointly are essential to feature alignment between the modalities. 
As noted in Table~\ref{tab:ablation_shots}, this tendency can also be observed for various shots - DAPT has superior performance compared with the combination of CoOp~\cite{zhou2022coop} and VPT~\cite{jia2022visual}. 

\begin{table*}[t]
\centering
\begin{tabular}{l c c c c c c}
\toprule
    Method & OxfordPets & Flowers102 & FGVCAircraft & DTD    & EuroSAT  & StanfordCars \\ 
    \midrule
CoOp+VPT                               & \textcolor[gray]{0.5}{91.90} & \textcolor[gray]{0.5}{96.89} & \textcolor[gray]{0.5}{46.06} & \textcolor[gray]{0.5}{69.86} & \textcolor[gray]{0.5}{91.77} & \textcolor[gray]{0.5}{82.78} \\
CoOp+VPT w/ $\mathcal{L}_\text{inter}$ & 91.97 & 96.85 & 46.52 & 70.06 & 92.01 & 82.95 \\
CoOp+VPT w/ $\mathcal{L}_\text{intra}$ & 91.97 & 97.03 & 45.90 & 70.76 & 92.16 & 83.14 \\
DAPT                                   & \textbf{92.27} & \textbf{97.06} & \textbf{46.37} & \textbf{71.38} & \textbf{92.65} & \textbf{83.03} \\ 
\midrule
Method & Food101 & SUN397 & Caltech101 & UCF101 & ImageNet & Average      \\ 
\midrule
CoOp+VPT                               & \textcolor[gray]{0.5}{86.52} & \textcolor[gray]{0.5}{75.88} & \textcolor[gray]{0.5}{95.70} & \textcolor[gray]{0.5}{84.23} & \textcolor[gray]{0.5}{72.14} & \textcolor[gray]{0.5}{81.25} \\
CoOp+VPT w/ $\mathcal{L}_\text{inter}$ & 86.42 & 75.71 & 95.71 & 84.27 & 72.07 & 81.33 \\
CoOp+VPT w/ $\mathcal{L}_\text{intra}$ & 86.47 & 75.90 & 95.74 & 84.35 & 72.15 & 81.41 \\
DAPT                                   & \textbf{86.55} & \textbf{75.99} & \textbf{95.82} & \textbf{84.53} & \textbf{72.20} & \textbf{81.62} \\
\bottomrule
\end{tabular}
\caption{\textbf{Abalation study.} We compared DAPT with the baseline, \textit{i.e.}, CoOp~\cite{zhou2022coop}+VPT~\cite{jia2022visual}, on 11 benchmark datasets. We observed performance gain in most datasets by adding losses one by one. }
\label{tab:ablation_loss}
\end{table*}

\noindent\textbf{Exploration of Intra-dispersion Loss. }
As discussed in Section~\ref{s:method}, the prototype $\boldsymbol{s}$ is defined as the average of image embeddings in DAPT. 
To evaluate the prototypes set by the mean of embeddings,
we conduct the ablation study of intra-dispersion loss with the prototype (DAPT-R) set by a random sample's embedding. 
Table~\ref{tab:ablation_uniform} shows the image classification results for 1, 2, 4, 8, and 16-shots on 11 datasets.
From the results, we confirmed that clustering around the average of samples (DAPT) is more effective than clustering around the random sample (DAPT-R),
especially when the number of shots is not extremely small (\textit{i.e.}, 4, 8, and 16-shots). 
\begin{table}[tbh]
\centering
\begin{tabular}{l c c c c c}
    \toprule
    \multirow{2}{*}[-0.2em]{Method} & \multicolumn{5}{c}{\# of training samples in each class} \\
    \cmidrule(l){2-6}
    & 1 & 2 & 4 & 8 & 16 \\
    \midrule
    CoOp+VPT & 61.05 & 68.49 & 73.28 & 78.76 & 81.25 \\
    DAPT-R & \textbf{65.08} & \textbf{70.51} & 74.85 & 78.11 & 81.29 \\
    DAPT & 61.42 & 69.95 & \textbf{74.91} & \textbf{78.98} & \textbf{81.62} \\
    \bottomrule
\end{tabular}
\caption{\textbf{Average performance on 11 datasets. } DAPT-R refers to applying intra-dispersion loss $\mathcal{L}_\text{intra}$ with the randomly defined prototype. }
\label{tab:ablation_uniform}
\end{table} 

\subsection{Analysis}
We investigate whether DAPT learns text and image embeddings as intended - spreading text embeddings and assembling image embeddings within the same classes - by quantitative and qualitative analyses. 

\noindent\textbf{Quantitative Analysis. }
\noindent We analyze the pairwise distance $\text{pdist}$ and the area of the convex hull of embeddings. 
Table~\ref{tab:numerical_pairwise_distance} shows the relative average pairwise distance $\text{pdist}$ between image embeddings of the same class compared to zero-shot CLIP, which is computed as the relative ratios, \ie, $\Delta \text{pdist (\%)}  = (1-\text{DAPT}/\text{zero-shot CLIP})\times 100$.
The results evidence that DAPT properly minimizes the average pairwise distance between image embeddings in a class ($\approx$ intra-dispersion)
and maximizes it for text (labels) embeddings across classes ($\approx$ inter-dispersion).
This implies that DAPT learns prompts to better latent spaces for feature alignment.

\begin{figure}[ht]
    \centering
    \begin{tabular}{ccc}
    \multirow{-8}{*}{\rotatebox[origin=c]{90}{Flowers102}} & \subfloat[Zero-shot CLIP]
    {\includegraphics[width=.18\textwidth]{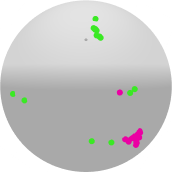}
    \label{fig:tsne_flower_zs}}
    & \subfloat[DAPT]{\includegraphics[width=.18\textwidth]{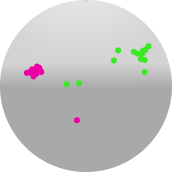}
    \label{fig:tsne_flower_dapt}}
    \vspace{0.5cm} \\
    \multirow{-8}{*}{\rotatebox[origin=c]{90}{UCF101}} & \subfloat[Zero-shot CLIP]{\includegraphics[width=.18\textwidth]{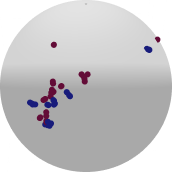}
    \label{fig:tsne_ucf_zs}} 
    & \subfloat[DAPT]{\includegraphics[width=.18\textwidth]{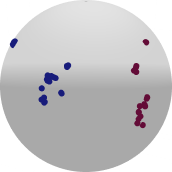}
    \label{fig:tsne_ucf_dapt}}  \\

    \end{tabular}
    \caption{
    \textbf{Visualization of image embeddings.} 
    We visualize image embeddings of Flowers102 and UCF101 via t-SNE. 
    Compared to zero-shot CLIP, our DAPT more compactly clusters image embeddings for each class.
    }
    \label{fig:qual}
\end{figure}

\begin{table}[h]
\centering
\begin{tabular}{lrrr}
\toprule
\multirow{2}{*}[-0.2em]{Dataset}& \multicolumn{1}{c}{Image} & \multicolumn{2}{c}{Text} \\ 
\cmidrule(r){2-2} \cmidrule{3-4}
\multicolumn{1}{c}{} & \multicolumn{1}{c}{$\Delta \text{pdist}$ (\%)} & \multicolumn{1}{c}{$\Delta \text{pdist}$ (\%)} & \multicolumn{1}{c}{$\text{cvx\_hull}_\text{text}$} \\ 
\midrule
OxfordPets & 0.2 & 23.4 & 65.8 \\
Flowers102 & -0.3 & 36.4 & 145.1 \\
FGVCAircraft & -9.3 & 62.5 & 246.4 \\
DTD & -6.4 & 119.5 & 713.8 \\
EuroSAT & -22.2 & 117.1 & 921.9 \\
StanfordCars & -2.4 & 30.6 & 87.8 \\
Food101 & -4.6 & 27.8 & 115.8 \\
SUN397 & -9.9 & 33.2 & 148.3 \\
Caltech101 & -2.8 & 45.2 & 115.3 \\
UCF101 & -1.4 & 51.1 & 251.4 \\
ImageNet & -0.3 & 0.2 & -12.2 \\ 
\bottomrule
\end{tabular}
\caption{
\textbf{Analysis of embeddings from DAPT.} Pairwise distances (pdist) and Convex hull area (cvx\_hull) of embeddings from DAPT compared to zero-shot CLIP. 
Ratios, $(1-\text{DAPT}/\text{zero-shot CLIP})\times 100$ are reported.
}
\label{tab:numerical_pairwise_distance}
\end{table}
\noindent\textbf{Qualitative Analysis. }
We verify that DAPT more compactly clusters image embeddings via t-SNE~\cite{van2008visualizing} visualization on several datasets. 
Figure~\ref{fig:qual} shows image embeddings of Fowers102~\cite{nilsback2008automated} and UCF101~\cite{soomro2012ucf101} benchmark datasets. 
Each point represents an image embedding, and the colors of the points indicate their classes. 
More t-SNE visualizations are provided in the supplement. 

\section{Limitations and Conclusion}
We proposed a distribution-aware prompt tuning method called DAPT for pre-trained vision-language models (VLMs). 
By considering the distribution of embeddings for prompt-tuning, which is underexplored in the literature, the proposed method significantly improves performance while maintaining the merits of existing prompt-tuning methods. 
In this paper, we present the inter-dispersion loss and intra-dispersion loss that appropriately optimize the text and visual latent spaces of VLMs, allowing us to achieve higher performance in downstream tasks using only prompts without additional layers. 
Although the proposed method significantly improves overall performance, it is still challenging to optimize prompts in the extreme few-shot settings, such as 1-shot and 2-shot. 
Lastly, it will be an interesting future direction to apply it to various downstream applications beyond image classification. 
\\
\section*{Acknowledgments}
This research was supported in part by the MSIT (Ministry of Science and ICT), Korea, under the ICT Creative Consilience program (IITP-2023-2020-0-01819) supervised by the IITP (Institute for Information \& communications Technology Planning \& Evaluation); the National Research Foundation of Korea (NRF) grant funded by the Korea government (MSIT) (NRF-2023R1A2C2005373); and the Google Cloud Research Credits program with the award  (MQMD-JNER-1YQC-YAQ5).

{\small
\bibliographystyle{ieee_fullname}
\bibliography{egbib}
}

\newpage
\appendix
\section{Implementation Details}
We implement DAPT based on the open source from CoOp~\cite{zhou2022coop} and VPT~\cite{jia2022visual}, using the PyTorch~\cite{paszke2017automatic} library. 
Before training, the learnable vectors for the text prompt are initialized with a zero-mean Gaussian distribution following the CoOp. In contrast, the learnable vectors for the visual prompt are initialized with the Xavier uniform initialization scheme following the VPT. In all experiments, the length of the learnable vector is set to 16 in both the text and visual prompt.
In the case of linear probe CLIP~\cite{radford2021learning} and zero-shot CLIP~\cite{radford2021learning}, we set the initialization of the text prompt as ``\texttt{a photo of a [CLASS].}" 
In the few-shot learning experiments, we follow the approach of Zhou~\etal~\cite{zhou2022coop} and conduct random sampling three times for each dataset. 
We report the average after testing three times for all experiments, including DAPT, CoOp~\cite{zhou2022coop}, VPT~\cite{jia2022visual}, and linear probe CLIP. 
As observed in the case of VPT, there is variance in the results of the visual prompt depending on hyperparameters such as learning rate. 
Therefore, we conduct a grid search for learning rate in the range of \{0.002, 0.02, 0.2, 2.0, 20.0\}, following the approach of Jia~\etal~\cite{jia2022visual}. 
\section{Additional Analyses}
In this section, we further analyze DAPT with various experiments.
\subsection{Analysis of Hyperparameter $\beta_t$ and $\beta_v$.}
\begin{figure}[h]
\centering
    \begin{tabular}{lr}
        \includegraphics[width=0.45\columnwidth]{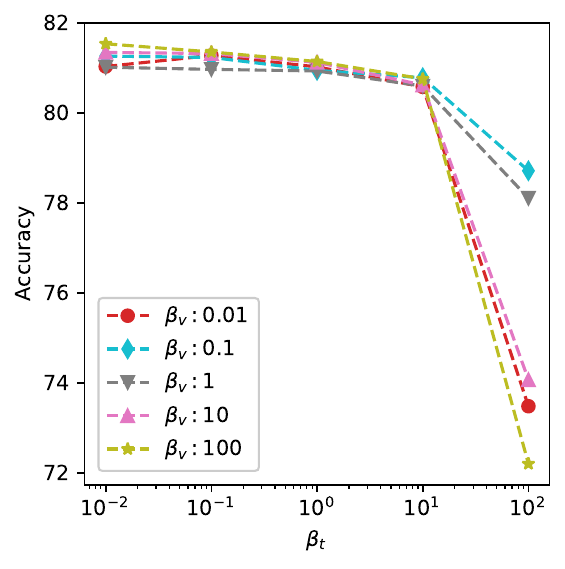} 
        &
        \includegraphics[width=0.45\columnwidth]{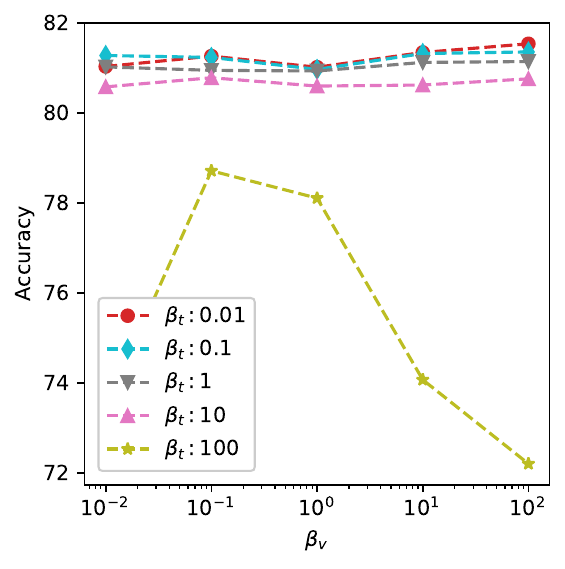}  \\
    \end{tabular}
\caption{Exploration of hyperparamters.}
\label{fig:supp_beta_sensitivity}
\end{figure}

The hyperparameter $\beta_t$ and $\beta_v$ adjust the strength of inter-dispersion loss and intra-dispersion loss, respectively. Due to the characteristics of each dataset, it may have different optimal values. We depict the accuracy according to $\beta_t$ and $\beta_v$ in Figure~\ref{fig:supp_beta_sensitivity} to investigate the hyperparameters. 

\noindent As described in Figure~\ref{fig:supp_beta_sensitivity}, DAPT shows consistent performance with a wide range of hyperparameters. 
To sum up, DAPT is robust to the choice of hyperparameters, except $\beta_t = 100$. 
Table~\ref{tab:supp_hyperparameter} summarizes optimal hyperparameters for each dataset.

\subsection{t-SNE Visualization}
\begin{figure*}
    \begin{tabular}{cccc}
    \multicolumn{2}{c}{(a) OxfordPets~\cite{parkhi2012cats}.} &
    \multicolumn{2}{c}{(b) Flowers102~\cite{nilsback2008automated}.} \\
    \vspace{0.1cm} \\
    \includegraphics[width=.23\linewidth]{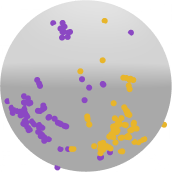} &
    \includegraphics[width=.23\linewidth]{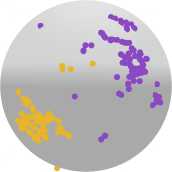} &
    \includegraphics[width=.23\linewidth]{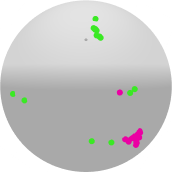} &
    \includegraphics[width=.23\linewidth]{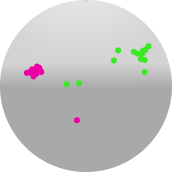} \\
    \vspace{0.1cm} \\
    
    \multicolumn{2}{c}{(c) DTD~\cite{cimpoi2014describing}.} &
    \multicolumn{2}{c}{(d) StanfordCars~\cite{krause20133d}.} \\
    \vspace{0.1cm} \\
    \includegraphics[width=.23\linewidth]{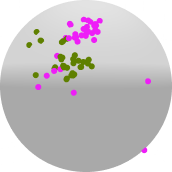} & 
    \includegraphics[width=.23\linewidth]{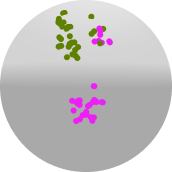} &
    \includegraphics[width=.23\linewidth]{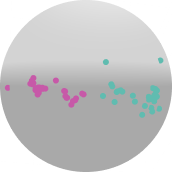} & 
    \includegraphics[width=.23\linewidth]{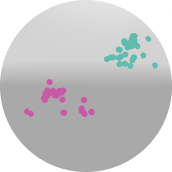} \\
    \vspace{0.1cm} \\
    
    \multicolumn{2}{c}{(e) FGVCAircraft~\cite{maji13fine-grained}.} &
    \multicolumn{2}{c}{(f) Caltech101~\cite{fei2004learning}.} \\
    \vspace{0.1cm} \\
    \includegraphics[width=.23\linewidth]{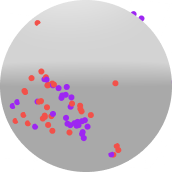} & 
    \includegraphics[width=.23\linewidth]{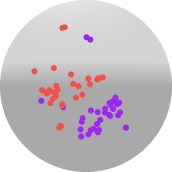} &
    \includegraphics[width=.23\linewidth]{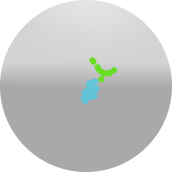} & 
    \includegraphics[width=.23\linewidth]{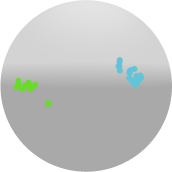} \\
    \vspace{0.1cm} \\

    \multicolumn{2}{c}{(e) UCF101~\cite{soomro2012ucf101}.} &
    \multicolumn{2}{c}{(i) EuroSAT~\cite{helber2019eurosat}.} \\
    \vspace{0.1cm} \\
    \includegraphics[width=.23\linewidth]{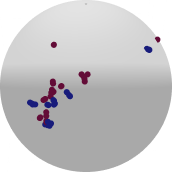} & 
    \includegraphics[width=.23\linewidth]{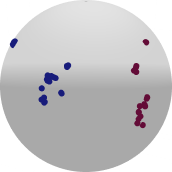} &
    \includegraphics[width=.23\linewidth]{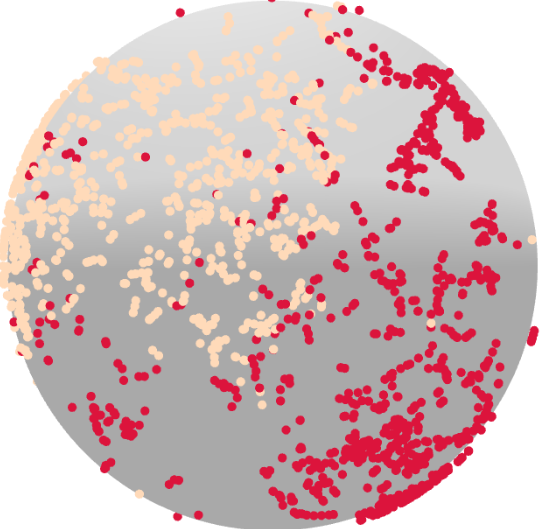} & 
    \includegraphics[width=.23\linewidth]{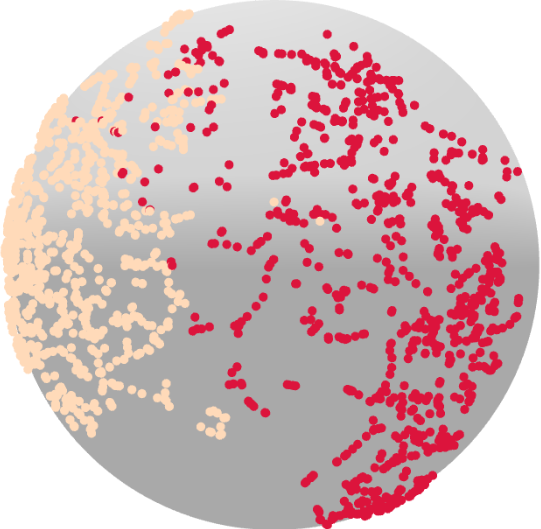} \\
    
    \end{tabular}
    \caption{\textbf{t-SNE~\cite{van2008visualizing} visualization of image embeddings.} In each dataset, the left hypersphere represents zero-shot CLIP, and the right hypersphere represents DAPT.}
    \label{fig:supple_tsne}
\end{figure*}

Figure~\ref{fig:supple_tsne} presents t-SNE~\cite{van2008visualizing} visualization of image embeddings in zero-shot CLIP~\cite{radford2021learning} and DAPT. 
All plots show that DAPT properly increases the distance between different classes as well as minimizes the intra-class variance.
Especially the results on OxfordPets~\cite{parkhi2012cats}, Flowers102~\cite{nilsback2008automated}, and UCF101~\cite{soomro2012ucf101} demonstrate that DAPT helps embeddings form compact clusters and increase the distance between different classes.
 
\subsection{Detailed Experimental Results}
In all experiments, we ran three times with randomly sampled data in each run and noted average values. For compelling results, we provide accuracy with standard deviation in 16-shots image classification on 11 datasets in Table~\ref{tab:rebuttal_mean_variance}. On average, DAPT achieved the best performance in 10 benchmarks.
\vspace{-2mm}
\begin{table}[h]
\centering
\footnotesize{
\begin{tabular}{l c c c c}
    \toprule
    Dataset & LP-CLIP & CoOp & VPT & DAPT \\
    \midrule
    OxfordPets & 86.49\stdev{0.06} & 91.91\stdev{0.42} & 92.04\stdev{0.58} & \textbf{92.27}\stdev{0.40} \\
    Flowers102 & \textbf{97.51}\stdev{0.13} & 96.79\stdev{0.35} & 91.48\stdev{0.15} & 97.06\stdev{0.25} \\
    FGVCAircraft & 45.51\stdev{0.08} & 43.96\stdev{0.74} & 34.92\stdev{0.16} & \textbf{46.37}\stdev{1.00} \\
    DTD & 69.58\stdev{0.73} & 69.98\stdev{0.18} & 61.47\stdev{0.37} & \textbf{71.38}\stdev{1.62} \\
    EuroSAT & 87.24\stdev{0.23} & 85.58\stdev{1.63} & 90.67\stdev{1.44} & \textbf{92.65}\stdev{0.86} \\
    StanfordCars & 80.67\stdev{0.46} & 82.62\stdev{0.13} & 70.59\stdev{1.00} & \textbf{83.03}\stdev{0.34} \\
    Food101 & 83.14\stdev{0.45} & 84.31\stdev{0.17} & 86.03\stdev{0.27} & \textbf{86.55}\stdev{0.10} \\
    SUN397 & 73.03\stdev{0.74} & 74.69\stdev{0.24} & 70.33\stdev{0.16} & \textbf{75.99}\stdev{0.12} \\
    Caltech101 & 95.51\stdev{0.31} & 95.68\stdev{0.20} & 95.35\stdev{0.15} & \textbf{95.82}\stdev{0.07} \\
    UCF101 & 82.35\stdev{0.36} & 82.15\stdev{1.42} & 79.99\stdev{0.69} & \textbf{84.53}\stdev{0.55} \\
    ImageNet & 67.42\stdev{0.26} & 71.93\stdev{0.10} & 69.31\stdev{0.05} & \textbf{72.20}\stdev{0.18} \\
    \bottomrule
\end{tabular}}
\vspace{-2mm}
\caption{16-shots image classification on 11 datasets.}
\vspace{-6mm}
\label{tab:rebuttal_mean_variance}
\end{table} 
\section{Generalization From Base to New Classes}
CoOp~\cite{zhou2022coop} demonstrated exemplary performance in the few-shot learning using text prompts, but it has a weak generalizability problem regarding unseen classes, as discussed in CoCoOp~\cite{zhou2022conditional}. As shown in Table 2, we prove that DAPT has significant performance gain in generalization. However, we supplement more experiments to prove that DAPT has superior generalization performance compared with baselines. In all experiments, we evaluate not only original classes but also unseen classes. Following Zhou~\etal~\cite{zhou2022conditional}, we divide the dataset into base classes and new classes, then train on 16 samples of the base class before testing on the new class. Similarly to the few-shot learning setting, we report the average of three times. The result for 11 datasets and the overall average is presented in Table~\ref{tab:supp_base2new}. The experimental results show that the accuracy for the new class is higher than CoOp in most datasets. The harmonic mean of the base and new class demonstrates superior performance for seven datasets.

\FloatBarrier
\FloatBarrier
\FloatBarrier
\begin{table*}[t]
\centering\resizebox{\textwidth}{!}{
\begin{tabular}{c c c}
    \begin{tabular}{lll|l}
    \multicolumn{4}{c}{\small{(a) Average over 11 datasets.}} \\
        \toprule
         & \multicolumn{1}{c}{Base} & \multicolumn{1}{c|}{New} & \multicolumn{1}{c}{H} \\
        \midrule
        CLIP & 69.53 & 74.34 & 71.85 \\
        CoOp & 82.71 & 62.84 & 71.42 \\
        DAPT & 84.20 & 63.71 & 72.54 \\
        \bottomrule
    \end{tabular} &
    \begin{tabular}{lll|l}
    \multicolumn{4}{c}{\small{(b) OxfordPets~\cite{parkhi2012cats}.}} \\
        \toprule
         & \multicolumn{1}{c}{Base} & \multicolumn{1}{c|}{New} & \multicolumn{1}{c}{H} \\
        \midrule
        CLIP & 91.33 & 97.15 & 94.15 \\
        CoOp & 93.37 & 95.43 & 94.39 \\
        DAPT & 94.00 & 72.43 & 81.82 \\
        \bottomrule
    \end{tabular} &
    \begin{tabular}{lll|l}
    \multicolumn{4}{c}{\small{(c) Flowers102~\cite{nilsback2008automated}.}} \\
        \toprule
         & \multicolumn{1}{c}{Base} & \multicolumn{1}{c|}{New} & \multicolumn{1}{c}{H} \\
        \midrule
        CLIP & 71.70 & 77.45 & 74.46 \\
        CoOp & 97.82 & 59.79 & 74.21 \\
        DAPT & 98.16 & 61.37 & 75.53 \\
        \bottomrule
    \end{tabular}  \\ \\

    \begin{tabular}{lll|l}
    \multicolumn{4}{c}{\small{(d) FGVCAircraft~\cite{maji13fine-grained}.}} \\
        \toprule
         & \multicolumn{1}{c}{Base} & \multicolumn{1}{c|}{New} & \multicolumn{1}{c}{H} \\
        \midrule
        CLIP & 27.67 & 35.87 & 31.24 \\
        CoOp & 40.66 & 24.44 & 30.53 \\
        DAPT & 45.54 & 19.74 & 27.54 \\
        \bottomrule
    \end{tabular} &
    \begin{tabular}{lll|l}
    \multicolumn{4}{c}{\small{(e) DTD~\cite{cimpoi2014describing}.}} \\
        \toprule
         & \multicolumn{1}{c}{Base} & \multicolumn{1}{c|}{New} & \multicolumn{1}{c}{H} \\
        \midrule
        CLIP & 53.24 & 60.87 & 56.80 \\
        CoOp & 79.59 & 40.30 & 53.51 \\
        DAPT & 82.06 & 53.42 & 64.71 \\
        \bottomrule
    \end{tabular} &
    \begin{tabular}{lll|l}
    \multicolumn{4}{c}{\small{(f) EuroSAT~\cite{helber2019eurosat}.}} \\
        \toprule
         & \multicolumn{1}{c}{Base} & \multicolumn{1}{c|}{New} & \multicolumn{1}{c}{H} \\
        \midrule
        CLIP & 56.93 & 63.92 & 60.22 \\
        CoOp & 92.21 & 50.70 & 65.43 \\
        DAPT & 95.05 & 43.02 & 59.23 \\
        \bottomrule
    \end{tabular} \\ \\
    
    \begin{tabular}{lll|l}
    \multicolumn{4}{c}{\small{(g) StanfordCars~\cite{krause20133d}.}} \\
        \toprule
         & \multicolumn{1}{c}{Base} & \multicolumn{1}{c|}{New} & \multicolumn{1}{c}{H} \\
        \midrule
        CLIP & 63.93 & 74.99 & 69.02 \\
        CoOp & 77.70 & 59.39 & 67.32 \\
        DAPT & 79.69 & 57.46 & 66.77 \\
        \bottomrule
    \end{tabular} &
    \begin{tabular}{lll|l}
    \multicolumn{4}{c}{\small{(h) Food101~\cite{bossard2014food}.}} \\
        \toprule
         & \multicolumn{1}{c}{Base} & \multicolumn{1}{c|}{New} & \multicolumn{1}{c}{H} \\
        \midrule
        CLIP & 90.08 & 91.13 & 90.60 \\
        CoOp & 88.40 & 85.87 & 87.11 \\
        DAPT & 89.57 & 89.82 & 89.69 \\
        \bottomrule
    \end{tabular} &
    \begin{tabular}{lll|l}
    \multicolumn{4}{c}{\small{(i) SUN397~\cite{xiao2010sun}.}} \\
        \toprule
         & \multicolumn{1}{c}{Base} & \multicolumn{1}{c|}{New} & \multicolumn{1}{c}{H} \\
        \midrule
        CLIP & 69.46 & 75.56 & 72.38 \\
        CoOp & 80.64 & 65.43 & 72.24 \\
        DAPT & 81.87 & 74.80 & 78.18 \\
        \bottomrule
    \end{tabular} \\ \\
    
    \begin{tabular}{lll|l}
    \multicolumn{4}{c}{\small{(j) Caltech101~\cite{fei2004learning}.}} \\
        \toprule
         & \multicolumn{1}{c}{Base} & \multicolumn{1}{c|}{New} & \multicolumn{1}{c}{H} \\
        \midrule
        CLIP & 97.22 & 94.21 & 95.69 \\
        CoOp & 98.19 & 86.17 & 91.79 \\
        DAPT & 98.24 & 87.74 & 92.69 \\
        \bottomrule
    \end{tabular} &
    \begin{tabular}{lll|l}
    \multicolumn{4}{c}{\small{(k) UCF101~\cite{soomro2012ucf101}.}} \\
        \toprule
         & \multicolumn{1}{c}{Base} & \multicolumn{1}{c|}{New} & \multicolumn{1}{c}{H} \\
        \midrule
        CLIP & 70.89 & 78.42 & 74.47 \\
        CoOp & 84.80 & 55.62 & 67.17 \\
        DAPT & 85.09 & 71.46 & 77.68 \\
        \bottomrule
    \end{tabular} &
    \begin{tabular}{lll|l}
    \multicolumn{4}{c}{\small{(l) ImageNet~\cite{deng2009imagenet}.}} \\
        \toprule
         & \multicolumn{1}{c}{Base} & \multicolumn{1}{c|}{New} & \multicolumn{1}{c}{H} \\
        \midrule
        CLIP & 72.40 & 68.12 & 70.19 \\
        CoOp & 76.44 & 68.11 & 72.04 \\
        DAPT & 76.97 & 69.54 & 73.07 \\
        \bottomrule
    \end{tabular} 
\end{tabular}}
\caption{Comparison of CLIP, CoOp, and DAPT in the base-to-new generalization setting.}
\label{tab:supp_base2new}
\end{table*}
\begin{table*}[t]
\centering
\begin{tabular}{lccccccccccc}
    \toprule
    Hyperparameters & \multicolumn{1}{c}{\rotatebox[origin=c]{90}{OxfordPets}} & \multicolumn{1}{c}{\rotatebox[origin=c]{90}{Flowers102}} & \multicolumn{1}{c}{\rotatebox[origin=c]{90}{FGVCAircraft}} & \multicolumn{1}{c}{\rotatebox[origin=c]{90}{DTD}} & \multicolumn{1}{c}{\rotatebox[origin=c]{90}{EuroSAT}} & \multicolumn{1}{c}{\rotatebox[origin=c]{90}{StanfordCars}} & \multicolumn{1}{c}{\rotatebox[origin=c]{90}{Food101}} & \multicolumn{1}{c}{\rotatebox[origin=c]{90}{SUN397}} & \multicolumn{1}{c}{\rotatebox[origin=c]{90}{Caltech101}} & \multicolumn{1}{c}{\rotatebox[origin=c]{90}{UCF101}} & \multicolumn{1}{c}{\rotatebox[origin=c]{90}{ImageNet}} \\ 
    \midrule
    
    $\beta_t$ & \multicolumn{1}{r}{0.1} & \multicolumn{1}{r}{0.01} & \multicolumn{1}{r}{0.01} & \multicolumn{1}{r}{0.01} & \multicolumn{1}{r}{10.0} & \multicolumn{1}{r}{0.1} & \multicolumn{1}{r}{0.01} & \multicolumn{1}{r}{0.01} & \multicolumn{1}{r}{0.01} & \multicolumn{1}{r}{0.1} & \multicolumn{1}{r}{0.01} \\
    
    $\beta_v$ & \multicolumn{1}{r}{10.0} & \multicolumn{1}{r}{10.0} & \multicolumn{1}{r}{100.0} & \multicolumn{1}{r}{100.0} & \multicolumn{1}{r}{100.0} & \multicolumn{1}{r}{100.0} & \multicolumn{1}{r}{10.0} & \multicolumn{1}{r}{100.0} & \multicolumn{1}{r}{10.0} & \multicolumn{1}{r}{0.1} & \multicolumn{1}{r}{0.01} \\

    Learning rate & \multicolumn{1}{r}{0.02} & \multicolumn{1}{r}{0.002} & \multicolumn{1}{r}{2.0} & \multicolumn{1}{r}{20.0} & \multicolumn{1}{r}{20.0} & \multicolumn{1}{r}{0.002} & \multicolumn{1}{r}{20.0} & \multicolumn{1}{r}{20.0} & \multicolumn{1}{r}{0.2} & \multicolumn{1}{r}{20.0} & \multicolumn{1}{r}{2.0} \\
    
    \bottomrule
\end{tabular}
\caption{Hyperparameters on 11 datasets in few-shot learning.}
\label{tab:supp_hyperparameter}
\end{table*}
\FloatBarrier 

\end{document}